\documentclass{article}

\usepackage{makecell}
\usepackage[final]{lg}

\usepackage[utf8]{inputenc} % allow utf-8 input
\usepackage[T1]{fontenc}    % use 8-bit T1 fonts
\usepackage{hyperref}       % hyperlinks
\usepackage{url}            % simple URL typesetting
\usepackage{booktabs}       % professional-quality tables
\usepackage{amsfonts}       % blackboard math symbols
\usepackage{nicefrac}       % compact symbols for 1/2, etc.
\usepackage{microtype}      % microtypography
\usepackage[dvipsnames]{xcolor}
\usepackage{kotex}
\usepackage{adjustbox}
\usepackage{multirow}
\usepackage{amssymb}
\usepackage{longtable}
\usepackage{comment}
\usepackage[autostyle]{csquotes}
\usepackage{hhline}
\usepackage{tabu}
\usepackage{tablefootnote}
\usepackage{longtable}
\usepackage{graphicx}
\usepackage{array}
\usepackage{caption}
\usepackage{tcolorbox}
\usepackage{lipsum}
\usepackage{relsize}
\usepackage{colortbl}
\usepackage{tablefootnote}
\tcbuselibrary{breakable}
\usepackage{floatpag}
\floatpagestyle{plain}
\usepackage{float}
\usepackage{threeparttable}

\usepackage{color}
\usepackage{soul}
\definecolor{lightlightgray}{rgb}{0.9,0.9,0.9}
\newcommand{\hlgrey}[1]{{\sethlcolor{lightlightgray} \hl{#1}}}

\usepackage{ulem}
\newcommand\kohlgrey[1][lightlightgray]{%
  \bgroup % 루아텍이면 주석처리
  \markoverwith{\textcolor{#1}{\vrule width.1em height.8em depth.2em}}%
  \ULon % 루아텍이면 주석처리
}

\usepackage{pifont}

\title{\centering EXAONE 3.5: \\ Series of Large Language Models for Real-world Use Cases}

\author{%
  LG AI Research\thanks{The complete list of authors who contributed to this work can be found in Appendix~\ref{appendix:contributors}.
}\\
}

\begin{document}

\maketitle
\addtocounter{footnote}{-1}

% section 0: ABSTRACT
\begin{abstract}

This technical report introduces the EXAONE 3.5 instruction-tuned language models, developed and released by LG AI Research. The EXAONE 3.5 language models are offered in three configurations: 32B, 7.8B, and 2.4B. These models feature several standout capabilities: 1) exceptional instruction following capabilities in real-world scenarios, achieving the highest scores across seven benchmarks, 2) outstanding long-context comprehension, attaining the top performance in four benchmarks, and 3) competitive results compared to state-of-the-art open models of similar sizes across nine general benchmarks. The EXAONE 3.5 language models are open to anyone for research purposes and can be downloaded from \url{https://huggingface.co/LGAI-EXAONE}. For commercial use, please reach out to the official contact point of LG AI Research: \href{mailto:contact_us@lgresearch.ai}{contact\_us@lgresearch.ai}.

\end{abstract}

% section 1: INTRODUCTION
\section{Introduction}

EXAONE 3.0 instruction-tuned large language model with 7.8B parameters~\citep{research2024exaone3078binstruction} demonstrated strong bilingual capabilities in Korean and English with exceptional real-world performance and instruction-following proficiency.
Since its release, we have received diverse feedback from both academic and industrial communities.
%Since the release of EXAONE 3.0 7.8B instruction-tuned language model~\citep{research2024exaone3078binstruction}, we have received diverse feedback from both academic and industrial communities.
For instance, academic researchers have emphasized the need for smaller models that can be trained and deployed on low-specification GPUs due to limited access to advanced computational infrastructure. The industry has expressed strong demand for larger models with enhanced performance that remain cost-effective, as well as smaller models suitable for on-device deployment. Additionally, with the increasing adoption of retrieval-augmented generation (RAG) techniques, which generate answers based on reference documents or web search results, there has been substantial demand for models capable of effectively handling longer contexts.

In this report, we present EXAONE 3.5 language models, a collection of instruction-tuned language models ranging from 2.4B to 32B parameters, developed to meet the diverse needs of users. EXAONE 3.5 language models include: 
1) 2.4B model optimized for deployment on small or resource-constrained devices, 
2) 7.8B model matching the size of its predecessor but offering improved performance, and 
3) 32B model delivering exceptional performance. 
All models support long-context processing of up to 32K tokens. Each model demonstrates state-of-the-art performance in real-world use cases and long-context handling, while remaining competitive in general domains compared to recently released models of similar sizes.

With the release of the EXAONE 3.5 language models, we hope to support researchers to push the boundaries of generative AI and inspire the development of innovative applications that enhance human life. This is in line with the mission of LG AI Research: \textsc{Advancing AI for a better Life}.

% section 2: MODEL TRAINING
\section{Model Training}

This section describes the detailed information on model configurations and the methods used for pre-training and post-training phases, along with the dataset construction process for each training phase.

\subsection{Model Configurations}

The EXAONE 3.5 language models are based on the latest decoder-only Transformer architecture, and detailed configurations are described in Table~\ref{tab:mod_arc}. These models are identical in structure to the EXAONE 3.0 7.8B model but mainly differ in their configurations related to sizes. Notably, the EXAONE 3.5 language models extend the maximum context length from 4,096 tokens in EXAONE 3.0 to 32,768 tokens by adopting the long-context fine-tuning~\citep{chen2023extendingcontextwindowlarge}. All three models share the same vocabulary, which consists roughly of 50\% Korean and 50\% English.

\begin{table}[t]
    \centering
    \small
    \setlength{\doublerulesep}{1pt}
    \begin{tabular}{l|ccc}
        \toprule
        Model size & 32B & 7.8B & 2.4B \\
        \midrule
        $d$\_model & 5,120 & 4,096 & 2,560  \\
        Number of layers & 64 & 32 & 30 \\
        Pre-normalization & True & True & True \\
        \midrule
        Non-linearity & SwiGLU \citep{shazeer2020gluvariantsimprovetransformer} & SwiGLU & SwiGLU \\
        Feedforward dimension & 27,392 & 14,336 & 7,168 \\
        \midrule
        Head type & GQA \citep{ainslie2023gqatraininggeneralizedmultiquery} & GQA & GQA \\
        Number of heads & 40 & 32 & 32 \\
        Number of KV heads & 8 & 8 & 8 \\
        Head size & 128 & 128 & 80 \\
        Max sequence length & 32,768 & 32,768 & 32,768 \\
        RoPE theta \citep{su2023roformerenhancedtransformerrotary} & 1,000,000 & 1,000,000 & 1,000,000 \\
        \midrule
        Tokenizer & BBPE \citep{wang2020bbpe} & BBPE & BBPE \\
        Vocab size & 102,400 & 102,400 & 102,400 \\
        Tied word embedding & False & False & True \\
        \bottomrule
    \end{tabular}
    \vspace{2mm}    
    \caption{Configurations of EXAONE 3.5 language models.}
    \label{tab:mod_arc}
\end{table}

\subsection{Pre-training}
\label{sec:pretraining}

The amount of pre-training corpus data and computational resources are shown in Table~\ref{tab:training_data_size}. The approach to data construction and model training consists of two stages: 1) we perform first-stage pre-training based on the large training corpus, which is collected and processed from as diverse sources as possible aimed to increase the performance on general domains. After that, 2) we collect more data for the domains that need to be strengthened through evaluations and conduct second-stage of pre-training. For instance, we focus on enhancing long-context understanding capabilities in the second-stage.

\begin{table}[htb!]
    \centering
    \small
    \setlength{\doublerulesep}{1pt}
    \begin{tabular}{l|ccc}
        \toprule
        Model size & 32B & 7.8B & 2.4B \\
        \midrule
        Training tokens & 6.5T & 9T & 6.5T \\
        Amount of computation (FLOPs) & $1.25 \times 10^{24}$ & $4.21 \times 10^{23}$ & $9.36 \times 10^{22}$ \\
        \bottomrule
    \end{tabular}
    \vspace{2mm}
    \caption{The sizes of the training data corpus along with the amounts of computation to build EXAONE 3.5 language models.}
    \label{tab:training_data_size}
\end{table}

\subsubsection{Context Length Extension}

To extend the context length, we utilize the long-context fine-tuning technique~\cite{chen2023extendingcontextwindowlarge}. To mitigate the catastrophic forgetting problem~\cite{MCCLOSKEY1989109}, where the model forgets what it learned during the first pre-training stage, a replay-based method~\citep{ai2024yiopenfoundationmodels} is applied. Specifically, during the second-stage pre-training, we reuse a portion of the data used in the first-stage. While documents exceeding the maximum context length is split into smaller chunks in the first-stage, the original corpus are trained without being divided into chunks in the second-stage to extend the models' context length.

\subsubsection{Decontamination}\label{subsec:decontam}

By the nature of massively web-crawled corpus, test-set examples often appear in the training corpus~\citep{riddell2024quantifying,xu2024benchmarking}.
These contaminated examples are likely to harm generalization performance and confuse test metrics, thus presenting unfair evaluations to users.
To prevent the contaminated examples undermine the generalization performance of EXAONE 3.5 language models, we rigorously apply a decontamination process for all targeted benchmark test data and remove contaminated examples from the training pipeline.

We borrow a simple yet powerful substring-level matching method~\citep{openai2023GPT4technicalreport} with stricter criteria.
The entire decontamination process is described in Figure~\ref{fig:decontam} in Appendix~\ref{appendix:decontam}.
We first normalize all test-set examples by removing all other characters except alphabets and numbers, then we extract all unique substrings with sliding window size $S=50$ and a stride of 1.
To determine whether a training example is contaminated, we randomly sample $N=10$ substrings from the normalized training example and check if they exist in the substring pools.
Table~\ref{tab:decon_example} in Appendix~\ref{appendix:decontam} provides examples of documents found in web corpora considered as contaminated.

\subsubsection{Training Cost}

Considering the computational cost of pre-training a large language model (LLM), it is necessary to make the training efficient by achieving as much high performance as possible with limited resources.
Table~\ref{tab:training_data_size_along_with_others} compares the total amounts of computations required for pre-training between the EXAONE 3.5 32B language model and others of similar size. When we simply approximate the total amounts of computations as the product of the model size and the number of training tokens~\cite{kaplan2020scalinglawsneurallanguage,hoffmann2022an}, Qwen 2.5 32B, for example, requires 2.77 times more computations than EXAONE 3.5 32B.
One of the noticeable characteristics of the EXAONE 3.5 language models is that they demonstrate high performance despite being trained at lower costs than the other baseline models (see Section~\ref{sec:eval}).

\begin{table}[htb!]
    \centering
    \setlength{\doublerulesep}{1pt}
    \begin{tabular}{c|c|c|c}
        \toprule
        Models & Model size & Training tokens & Amount of computation (ratio) \\
        \midrule
        EXAONE 3.5 & 32B & 6.5T    & 1.00 \\
        Qwen 2.5   & 32B & 18T     & 2.77 \\
        Gemma 2    & 27B & 13T     & 1.69 \\
        Yi 1.5     & 34B & 3.6T    & 0.59 \\
        \bottomrule
    \end{tabular}
    \vspace{2mm}
    \caption{Comparison of the total amounts of computations to build models. We approximate the amount of computations as the product of the model size and the number of training tokens. Although the EXAONE 3.5 32B model is behind in the computations compared to Qwen 2.5 and Gemma 2, it has shown competitive performances.}
    \label{tab:training_data_size_along_with_others}
\end{table}

\subsection{Post-training}

After pre-training, models go through further processes for strengthening their instruction-following capabilities and aligning with human preferences, which are well known as supervised fine-tuning (SFT)~\citep{wei2022finetunedlanguagemodelszeroshot} and preference optimization.

\subsubsection{Supervised Fine-tuning}

\begin{figure}[htbp]
    \centering
    \includegraphics[scale=0.3]{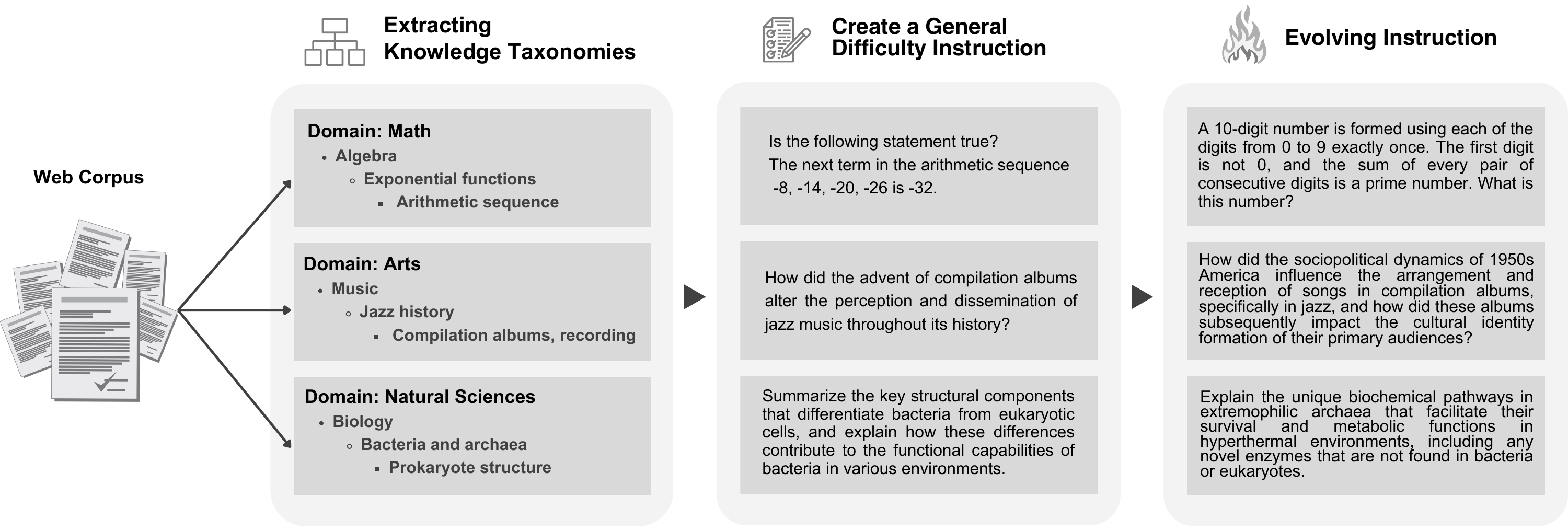}
    \caption{A procedure of instruction-tuning data construction. First, we extract the core knowledge from large-volume web corpora and classify it within the taxonomy we defined in advance. Next, instruction-tuning data is generated based on the knowledge. To construct additional training data that is more complex, we leverage an instruction-evolving method~\citep{zeng-etal-2024-automatic} that lets the final dataset cover various fields with varying levels of difficulty.
    }
    
    \label{fig:sft_pipeline}
\end{figure}

To perform well on new or unseen instructions, a model needs to be trained on pairs of instruction-response datasets with varying difficulty from different domains. Hence, in order to build training data covering a wide range of fields, we extract core knowledge from 8M web corpora using a taxonomic system, as shown in Figure~\ref{fig:sft_pipeline}. We then generate an instruction-tuning dataset based on the extracted knowledge taxonomy. Finally, leveraging an instruction evolution method, which stems from the method proposed in~\citep{zeng-etal-2024-automatic}, we diversify the complexity levels so that instructions with various complexities and difficulties can be produced.

\subsubsection{Preference Optimization}

\begin{figure}[htbp]
    \centering
    \includegraphics[scale=0.95]{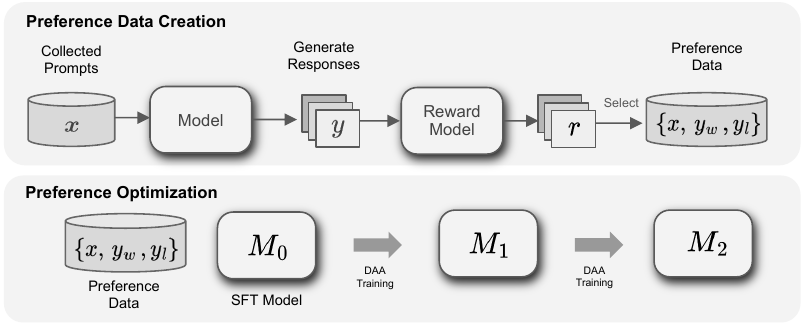}
    \caption{Overview of the preference optimization pipeline. (Top) Preference Data Creation: It shows the process of constructing preference data $\{x, y_w, y_l\}$ by scoring the responses $y$ generated from a model for the prompt $x$ using a reward model. (Bottom) Preference Optimization: Sequential training process where $M_0$ initialized from the SFT model is trained through DAA to obtain $M_1$ and $M_2$.}
    \label{fig:dpo_pipeline}
\end{figure}

Direct alignment algorithms (DAAs)~\cite{rafailov2024scaling}, such as DPO~\citep{rafailov2023directpreferenceoptimizationlanguage} and SimPO~\citep{meng2024simposimplepreferenceoptimization}, are used to train models after supervised fine-tuning to align models with human preferences. We create preference data for training using synthetic data and pre-collected data. For response generation, we sample $N$ responses from multiple models for the prompt $x$ drawn from the preference data and select the best response as $y_w$ and the worst response as $y_l$ based on the scores of a reward model to create a preference data, $\{x, y_w, y_l\}$. To validate preference data, we use an additional reward model to calculate agreement based on the rankings of the two reward models and filter out data with agreement below the threshold. Our preference optimization comprises multiple stages to sequentially train models $M_1$ and $M_2$ through DAAs, where $M_0$ is initialized from the SFT model. The staged pipeline enables us to mitigate over-optimization~\citep{rafailov2024scaling} that may occur during the DAAs’ training process. Figure~\ref{fig:dpo_pipeline} shows a schematic diagram for constructing our preference dataset and training process.

\subsection{Data Compliance}
\label{sec:data_compliance}

Developing AI models requires a large amount of data, and the acquisition and utilization of this data can lead to various legal issues, such as copyright infringement, intellectual property infringement, and personal information protection violations. To minimize these risks, LG AI Research conducts AI Compliance reviews throughout the entire process of data collection, AI model training, and information provision. For more detailed information, please refer to the EXAONE 3.0 Technical Report~\citep{research2024exaone3078binstruction} and the LG AI Ethics Principles~\citep{lgethics}.

% section 3: EVALUATION
\section{Evaluation}\label{sec:eval}

This section presents the evaluation settings and results of EXAONE 3.5 language models on various benchmark datasets. We select recently released open language models for baselines of our models to compare our performances on the benchmarks. All baselines and their detailed information are described in Appendix~\ref{appendix:baseline_details}.

\subsection{Benchmarks}
\label{subsec:benchmarks}

Considering the diverse nature of user intents, it is crucial for an instruction-tuned model to generate a response aligned to the user's query, whatever it is. 
To evaluate our models in comprehensive and various scenarios, we select over a dozen evaluating benchmarks along with a few in-house benchmarks.
Table~\ref{tab:benchmark_set} summarizes all benchmarks, which can be grouped into three categories:

\begin{itemize}
    \item \textbf{Real-world Use Cases} (Section~\ref{subsec:real_bench}): the benchmarks requiring the ability to understand and perform diverse user instructions.
    \item \textbf{Long Context} (Section~\ref{subsec:long_bench}): the benchmarks evaluating the ability to understand the long context.
    \item \textbf{General Domain} (Section~\ref{subsec:general_bench}): the benchmarks embracing general domain abilities that LLMs are expected to have. Specifically, this category includes benchmarks for measuring the ability to solve mathematical problems, the ability to write source codes, and the parametric knowledge embedded in an LLM.
\end{itemize}

\begin{table}[h!]
    \small
    \centering
    \setlength{\doublerulesep}{1pt}
    \begin{tabular}{l|l|c|l|c}
        \toprule
        Category & Benchmark & Lang & Evaluation Settings & Metric \\
        \midrule
        Real-world Use Cases & MT-Bench \cite{zheng2023judging} & EN & LLM-as-a-judge{\smaller[3]~(judge: \textsl{gpt-4o-2024-08-06})}\tablefootnote{The separability of the original GPT-4 judge results is notably low, prompting the adoption of \textsl{gpt-4o-2024-08-06} as judge.} & LLM score \\
        & LiveBench \citep{white2024livebenchchallengingcontaminationfreellm}{\smaller[3]~(v2024-08-31)} & EN & Ground-truth match & Accuracy \\
        & Arena-Hard-v0.1 \citep{li2024crowdsourceddatahighqualitybenchmarks} & EN & LLM-as-a-judge{\smaller[3]~(judge: \textsl{gpt-4-1106-preview})} & Win rate \\
        & AlpacaEval 2.0 LC \citep{dubois2024lengthcontrolledalpacaevalsimpleway} & EN & LLM-as-a-judge{\smaller[3]~(judge: \textsl{gpt-4-1106-preview})} & Win rate \\
        & IFEval\citep{zhou2023instructionfollowingevaluationlargelanguage} & EN & Prompt-level / strict accuracy & Accuracy \\
        & KoMT-Bench \citep{komt-bench} & KO & LLM-as-a-judge{\smaller[3]~(judge: \textsl{gpt-4o-2024-08-06})} & LLM score \\
        & LogicKor \citep{logickor} & KO & LLM-as-a-judge{\smaller[3]~(judge: \textsl{gpt-4-1106-preview})} & LLM score \\
        \midrule
        Long Context & Needle-In-A-Haystack \citep{kamradt2023llmtest} & EN/KO & Ground-truth match & Accuracy \\
        & LongBench \citep{bai-etal-2024-longbench} & EN & Ground-truth match & F1, Rouge \\
        & LongRAG~\citep{jiang2024longragenhancingretrievalaugmentedgeneration}{\smaller[3]~(extended)} & EN & LLM-as-a-judge{\smaller[3]~(judge: \textsl{gpt-4o-2024-08-06})} & LLM score \\
        & Ko-LongRAG{\smaller[3]~(In-house)} & KO & LLM-as-a-judge{\smaller[3]~(judge: \textsl{gpt-4o-2024-08-06})} & LLM score \\
        & Ko-WebRAG{\smaller[3]~(In-house)} & KO & LLM-as-a-judge{\smaller[3]~(judge: \textsl{gpt-4o-2024-08-06})} & LLM score \\
        \midrule
        General Domain & GSM8K \citep{cobbe2021trainingverifierssolvemath} & EN & 0-shot / CoT & Accuracy \\
        & MATH \citep{hendrycks2021measuringmathematicalproblemsolving, minerva_math} & EN & 0-shot / CoT & Accuracy \\
        & HumanEval \citep{chen2021evaluatinglargelanguagemodels} & EN & 0-shot & pass@1 \\
        & MBPP \citep{austin2021programsynthesislargelanguage} & EN & 0-shot{\smaller[3]~(Evalplus base)\tablefootnote{We choose the \textsc{MBPP} base from EvalPlus~\cite{liu2023is}, which is a subset of the original and consists of refined, high-quality problems.}} & pass@1 \\
        & GPQA \citep{rein2023gpqagraduatelevelgoogleproofqa} & EN & 0-shot / CoT & Accuracy \\
        & ARC-C \citep{clark2018thinksolvedquestionanswering} & EN & 0-shot & Accuracy \\
        & BBH \citep{suzgun2022challenging} & EN & 0-shot / CoT & Accuracy \\
        & MMLU \citep{hendrycks2020measuring} & EN & 0-shot / CoT & Accuracy \\
        & KMMLU \citep{son2024kmmlumeasuringmassivemultitask} & KO & 0-shot / CoT & Accuracy \\
        \bottomrule
    \end{tabular}
    \vspace{2mm}
    \caption{The benchmarks used to evaluate the performance of EXAONE 3.5 language models along with their target languages, evaluation settings, and the metrics. \textsc{LongRAG} is extended from the original, and \textsc{Ko-LongRAG} and \textsc{Ko-WebRAG} are in-house benchmarks (see Section~\ref{sec:long_context}).}
    \label{tab:benchmark_set}
\end{table}

\subsection{Overall Performance}\label{subsec:overall_result}

The results of overall performance against three categories are presented in Table~\ref{tab:eval_overall}. Our EXAONE 3.5 language models, with sizes 32B and 7.8B, perform best in Real-world Use Cases and Long Context categories compared to baseline models while showing competitive results in the General Domain category. Our smallest model, EXAONE 3.5 2.4B, outperforms baselines with similar sizes in all three categories, demonstrating strong performance.
Surprisingly, our 2.4B model, despite its small size, has shown better performance compared to baselines even with a larger size (<~9B) except for Qwen 2.5 7B in General Domain. Considering the recent surge in demand for smaller large language models (sLLM)~\citep{sllmsurvey2024}, we believe that our EXAONE 3.5 2.4B model is well-positioned to be highly competitive in both academic and industrial use.

In the following sections, we elaborate on detailed evaluation settings and the results for each category.

\begin{table}[t!]
\tabcolsep=0.12cm
    \small
    \centering
    \begin{tabular}{wc{0.245\linewidth}|wc{0.185\linewidth}wc{0.185\linewidth}wc{0.185\linewidth}}
        \toprule
        Models & Real-world Use Cases & Long Context & General Domain \\
        \midrule
        \rowcolor[rgb]{0.9,0.9,0.9} EXAONE 3.5 32B
        & \textbf{74.3} & \textbf{71.1} & \underline{74.8} \\
        Qwen 2.5 32B~\cite{qwen2.5}
        & \underline{69.8} & \underline{66.9} & \textbf{78.7} \\
        C4AI Command R 32B
        ~\cite{cohere_for_ai_2024}
        & 46.0 & 63.4 & 56.8 \\
        Gemma 2 27B~\cite{gemmateam2024gemma2improvingopen}
        & 64.2 & - & 68.7 \\
        Yi 1.5 34B~\cite{ai2024yiopenfoundationmodels}
        & 46.9 & - & 53.9 \\
        \midrule
        \rowcolor[rgb]{0.9,0.9,0.9} EXAONE 3.5 7.8B
        & \textbf{70.7} & \textbf{66.6} & \underline{70.2} \\
        Qwen 2.5 7B~\cite{qwen2.5}
        & 52.7 & 56.1 & \textbf{71.0} \\
        Llama 3.1 8B~\cite{grattafiori2024llama3herdmodels}
        & 48.6 & \underline{58.8} & 62.4 \\
        Gemma 2 9B~\cite{gemmateam2024gemma2improvingopen}
        & \underline{57.9} & - & 62.9 \\
        Phi 3 small (7B)~\cite{abdin2024phi3technicalreporthighly}
        & 41.7 & 33.4 & 63.2 \\
        \midrule
        \rowcolor[rgb]{0.9,0.9,0.9} EXAONE 3.5 2.4B
        & \textbf{61.1} & \textbf{63.4} & \textbf{63.3} \\
        Qwen 2.5 3B~\cite{qwen2.5}
        & \underline{44.5} & 40.7 & \underline{62.1} \\
        Qwen 2.5 1.5B~\cite{qwen2.5}
        & 30.1 & 34.5 & 47.9 \\
        Llama 3.2 3B~\cite{llama3.2}
        & 36.7 & \underline{44.2} & 54.9 \\
        Gemma 2 2B~\cite{gemmateam2024gemma2improvingopen}
        & 41.7 & - & 42.2 \\
        \bottomrule
    \end{tabular}
    \vspace{2mm}
    \caption{Overall comparison results of EXAONE 3.5 language models with similar-sized baseline language models. Here, a dash (-) indicates the model does not support context lengths longer than 16K. \textbf{Bold} scores indicate the best performance, and \underline{underlined} scores mean the second best. The detailed information for each baseline is described in Appendix~\ref{appendix:baseline_details}.}
    \label{tab:eval_overall}
\end{table}

\subsection{Real-world Use Cases} \label{subsec:real_bench}

For the Real-world Use Cases category, we have compiled seven benchmarks that represent real-world queries users might submit to a chatbot model. In \textsc{MT-Bench}, \textsc{KoMT-Bench}, and \textsc{LogicKor}, models' responses consisting of multi-turns are evaluated by a judge model. For \textsc{Arena-Hard} and \textsc{AlpacaEval}, responses of a target language model are compared with those of a reference model (\textsl{gpt-4-0314} and \textsl{gpt-4-1106-preview}, respectively) by a judge model, recording the win rate. \textsc{LiveBench} (ver. 2024-08-31) and \textsc{IFEval} (prompt-strict) assess how well the models' responses align with user instructions by matching them to the ground-truth responses.

\begin{table}[htb!]
    \tabcolsep=0.12cm
    \small
    \centering
    \begin{tabular}{wc{0.185\linewidth}|wc{0.082\linewidth}wc{0.082\linewidth}wc{0.082\linewidth}wc{0.082\linewidth}wc{0.082\linewidth}wc{0.082\linewidth}wc{0.082\linewidth}|wc{0.082\linewidth}}
        \toprule
        Models & MT-Bench & LiveBench & Arena-Hard & AlpacaEval & IFEval & KoMT-Bench & LogicKor & Average \\
        \midrule
        \rowcolor[rgb]{0.9,0.9,0.9} EXAONE 3.5 32B & \textbf{8.51} & \underline{43.0} & \textbf{78.6} & \textbf{60.6} & \textbf{81.7} & \textbf{8.05} & \textbf{9.06} & \textbf{74.3} \\
        Qwen 2.5 32B   & \underline{8.49} & \textbf{50.6} & \underline{67.0} & 41.0 & \underline{78.7} & \underline{7.75} & \underline{8.89} & \underline{69.8} \\
        C4AI Command R 32B  & 7.38 & 29.7 & 17.0 & 25.9 & 26.1 & 6.72 & 8.24 & 46.0 \\
        Gemma 2 27B    & 8.28 & 40.0 & 57.5 & \underline{52.2} & 59.7 & 7.19 & 8.56 & 64.2 \\
        Yi 1.5 34B     & 7.64 & 26.2 & 23.1 & 34.8 & 55.5 & 4.88 & 6.33 & 46.9 \\
        \midrule
        \rowcolor[rgb]{0.9,0.9,0.9} EXAONE 3.5 7.8B & \textbf{8.29} & \textbf{39.8} & \textbf{68.7} & \textbf{54.2} & \textbf{78.9} & \textbf{7.96} & \textbf{9.08} & \textbf{70.7} \\
        Qwen 2.5 7B     & 6.48 & \underline{35.6} & \underline{48.9} & 31.7 & 72.5 & 5.19 & 6.38 & 52.7 \\
        Llama 3.1 8B    & 7.59 & 28.3 & 27.7 & 25.7 & \underline{74.5} & 4.85 & 5.99 & 48.6 \\
        Gemma 2 9B      & \underline{7.64} & 32.1 & 43.6 & \underline{47.3} & 54.7 & \underline{7.10} & \underline{8.05} & \underline{57.9} \\
        Phi 3 small (7B)        & 7.63 & 27.9 & 26.8 & 29.2 & 59.5 & 3.22 & 3.99 & 41.7 \\
        \midrule
        \rowcolor[rgb]{0.9,0.9,0.9} EXAONE 3.5 2.4B & \textbf{7.81} & \textbf{33.0} & \textbf{48.2} & \textbf{37.1} & \textbf{73.6} & \textbf{7.24} & \textbf{8.51} & \textbf{61.1} \\
        Qwen 2.5 3B     & \underline{7.21} & \underline{25.7} & \underline{26.4} & 17.4 & 60.8 & \underline{5.68} & 5.21 & \underline{44.5} \\
        Qwen 2.5 1.5B   & 5.72 & 19.2 & 10.6 & 8.4 & 40.7 & 3.87 & 3.60 & 30.1 \\
        Llama 3.2 3B  & 6.94 & 24.0 & 14.2 & 18.7 & \underline{70.1} & 3.16 & 2.86 & 36.7 \\
        Gemma 2 2B    & 7.20 & 20.0 & 19.1 & \underline{29.1} & 50.5 & 4.83 & \underline{5.29} & 41.7 \\
        \bottomrule
    \end{tabular}
    \vspace{2mm}
    \caption{Performance comparison results of EXAONE 3.5 language models with similar-sized recently-released language models on seven benchmarks representing real-world use case scenarios. When calculating the macro average, the scores of MT-Bench, KoMT-Bench, and LogicKor are multiplied by 10 because they are scored out of 10 and the rest are scored out of 100. \textbf{Bold} scores indicate the best performance, and \underline{underlined} scores mean the second best.}
    \label{tab:eval_real}
\end{table}

As presented in Table~\ref{tab:eval_real}, our three models have shown the best performance against baselines of similar size in all benchmarks, except for the 32B model in \textsc{LiveBench}. Furthermore, by outperforming others in both English and Korean benchmarks, EXAONE 3.5 language models demonstrate their superior bilingual abilities.

\subsection{Long Context} \label{subsec:long_bench}
\label{sec:long_context}

The ability to process and understand long contexts is increasingly important for modern LLMs, as it enables their application in more complex scenarios. To demonstrate EXAONE 3.5 language models' long context performance, we evaluate our models using benchmarks designed for a synthetic task with long context inputs, along with various retrieval-augmented generation (RAG) benchmarks.

\subsubsection{Needle-in-a-Haystack}
Needle-in-a-Haystack (NIAH) \citep{kamradt2023llmtest} serves as a benchmark to assess how effectively models can locate and retrieve information hidden at random locations within long documents. We comprehensively evaluate our models' ability to process and retrieve information from long contexts, up to 32K tokens. Furthermore, we extend NIAH to Korean and employ it to evaluate our models' long context processing ability across both English and Korean contexts.

\begin{figure}[t!]
    \centering
    \includegraphics[width=0.7\textwidth]{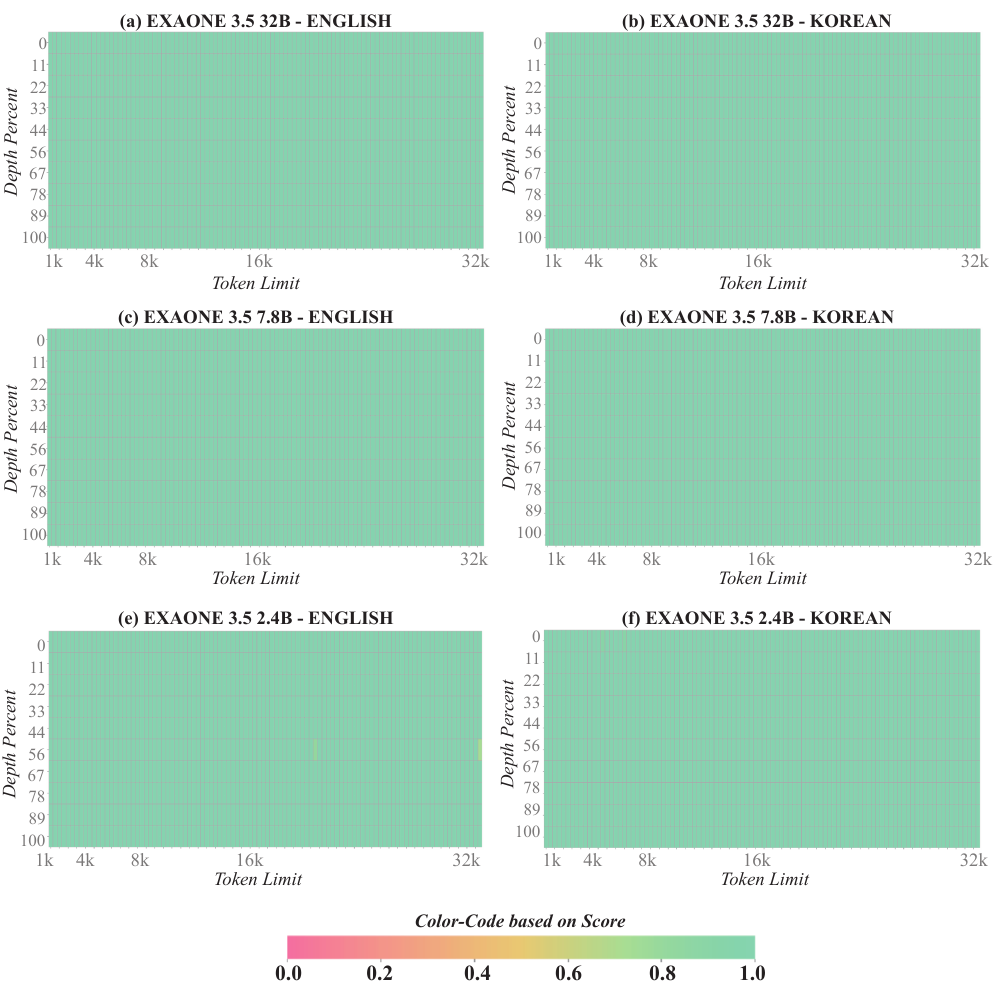}
    \vspace{5mm} 
    \caption{NIAH results of EXAONE 3.5 language models. The x-axis represents the token length of the input text, while the y-axis shows the relative position within the text, expressed as a percentage (0\% corresponds to the beginning, and 100\% to the end). The results are represented using a color-coded scheme: green indicates successful retrievals, and red represents unsuccessful ones. EXAONE 3.5 language models achieve near-perfect accuracy in retrieving information across various document depths and context lengths in English and Korean.}
    \label{fig:NIAH}
\end{figure}

Figure~\ref{fig:NIAH} demonstrates that our models achieve near-perfect accuracy in retrieving targeted information across all tested document depths and context lengths in both English and Korean. These results highlight their robust long context processing capabilities, particularly in tasks demanding precise information retrieval and complex reasoning.

\subsubsection{Long Context Understanding}

To assess long context understanding capabilities, we evaluate our models using benchmarks including \textsc{LongBench}~\citep{bai-etal-2024-longbench} and \textsc{LongRAG}~\citep{jiang2024longragenhancingretrievalaugmentedgeneration}.
We expand unanswerable cases in LongRAG to make it more challenging.
We also build \textsc{Ko-LongRAG}, the Korean counterpart to \textsc{LongRAG}, to evaluate long context understanding in Korean.
For more realistic RAG scenario, requiring answers to difficult questions using actual web-searched results, we constructed \textsc{Ko-WebRAG} benchmark.
We refer readers to the Appendix~\ref{appendix:lc_setting} for more details.

As shown in Table~\ref{tab:eval_long}, EXAONE 3.5 language models have shown superior performance compared to other models\footnote{Gemma models and Yi 1.5 34B model are excluded from evaluations due to their context limits ($\le$ 16k tokens), ensuring fair comparison.}, except for the 32B and 7.8B models in LongBench. When averaged across the benchmarks, our three models outperform all baselines, confirming their capabilities to process complex, extended contexts effectively.

\begin{table}[tb!]
    \tabcolsep=0.12cm
    \small
    \centering
    \begin{tabular}{wc{0.20\linewidth}|wc{0.12\linewidth}wc{0.12\linewidth}wc{0.12\linewidth}wc{0.12\linewidth}|wc{0.12\linewidth}}
        \toprule
        Models & LongBench & LongRAG & Ko-LongRAG & Ko-WebRAG & Average \\
        \midrule
        \rowcolor[rgb]{0.9,0.9,0.9} EXAONE 3.5 32B & \underline{49.2} & \textbf{67.6} & \textbf{85.3} & \textbf{82.3} & \textbf{71.1} \\
        Qwen 2.5 32B   & 49.1 & \underline{63.6} & \underline{73.5} & \underline{81.3} & \underline{66.9} \\
        C4AI Command R 32B  & \textbf{50.9} & 55.3 & 72.3 & 75.0 & 63.4 \\
        Gemma 2 27B    & - & - & - & - & - \\
        Yi 1.5 34B     & - & - & - & - & - \\
        \midrule
        \rowcolor[rgb]{0.9,0.9,0.9} EXAONE 3.5 7.8B & \underline{46.0} & \textbf{68.3} & \textbf{71.7} & \textbf{80.3} & \textbf{66.6} \\
        Qwen 2.5 7B     & \textbf{47.2} & \underline{60.1} & 55.3 & 61.7 & 56.1 \\
        Llama 3.1 8B    & 44.6 & 55.1 & \underline{64.8} & \underline{70.7} & \underline{58.8} \\
        Gemma 2 9B      & -        & -        & -        & -        & -        \\
        Phi 3 small (7B)        & 40.6 & 52.7 &  7.7 & 32.7 & 33.4 \\
        \midrule
        \rowcolor[rgb]{0.9,0.9,0.9} EXAONE 3.5 2.4B & \textbf{42.7} & \textbf{63.3} & \textbf{74.7} & \textbf{73.0} & \textbf{63.4} \\
        Qwen 2.5 3B     & \underline{42.0} & 45.8 & \underline{40.5} & 34.7& 40.7 \\
        Qwen 2.5 1.5B   & 37.1 & 39.0 & 33.8 & 28.0 & 34.5 \\
        Llama 3.2 3B    & 41.7 & \underline{45.9} & 39.3 & \underline{50.0} & \underline{44.2} \\
        Gemma 2 2B      & - & - & - & - & - \\
        \bottomrule
    \end{tabular}
    \vspace{2mm}
    \caption{Performance comparison results of EXAONE 3.5 language models with similar-sized recently released language models across four benchmarks representing long context scenarios. A dash (-) indicates that the model does not support context lengths longer than 16K. Context lengths for each model are detailed in Table~\ref{tab:baselines}. The average score in the rightmost is calculated as a macro average across the benchmarks. \textbf{Bold} scores indicate the best performance, and \underline{underlined} scores mean the second best.}
    \label{tab:eval_long}
\end{table}

\subsection{General Domain} \label{subsec:general_bench}

Language models are now expected to achieve human-level capabilities in various general domains, such as solving mathematical problems or writing source code programs. To evaluate overall performance in the general domains, we select nine benchmarks in three main domains: 1) \textsc{GSM8K} (CoT) and \textsc{MATH} (CoT) for mathematics, 2) \textsc{HumanEval} (Evalplus base) and \textsc{MBPP} (Evalplus base) for coding, and 3) \textsc{MMLU} (CoT), \textsc{KMMLU} (CoT), \textsc{GPQA} (CoT), \textsc{ARC-C}, and \textsc{BBH} (CoT) for assessing the amount of knowledge embedded in an LLM.

To better simulate the real-world scenarios where a chatbot model usually receives a single query from users, we evaluate all benchmarks in the General Domain category using the \textsl{0-shot setting}. To achieve this, we prompt language models with instructions that require specific answer formats and parse the final answer from the responses. For a fair comparison, we use the same prompts across all models. We make public all the prompts we used in Appendix~\ref{appendix:general_dom_setting} for transparent reproducibility.

\begin{table}[htb]
    \tabcolsep=0.12cm
    \small
    \centering
    \begin{tabular}{wc{0.175\linewidth}|wc{0.065\linewidth}wc{0.065\linewidth}wc{0.065\linewidth}wc{0.065\linewidth}wc{0.065\linewidth}wc{0.065\linewidth}wc{0.065\linewidth}wc{0.065\linewidth}wc{0.065\linewidth}|wc{0.065\linewidth}}
        \toprule
        Models & GSM8K & MATH & HumanEval & MBPP & MMLU & KMMLU & GPQA & ARC-C & BBH & Average \\
        \midrule
        \rowcolor[rgb]{0.9,0.9,0.9} EXAONE 3.5 32B & \underline{91.9} & \underline{70.5} & \underline{87.2} & \underline{81.8} & \underline{78.3} & \underline{57.0} & \underline{39.7} & 91.7 & \underline{75.3} & \underline{74.8} \\
        Qwen 2.5 32B   & \textbf{92.0} & \textbf{76.5} & \textbf{89.0} & \textbf{88.9} & \textbf{81.4} & \textbf{62.1} & \textbf{40.9} & \textbf{95.1} & \textbf{82.7} & \textbf{78.7} \\
        C4AI Command R 32B  & 56.5 & 24.3 & 68.3 & 78.8 & 71.1 & 41.5 & 27.4 & 88.0 & 55.7 & 56.8 \\
        Gemma 2 27B    & 84.2 & 49.4 & 79.3 & 80.7 & 74.8 & 53.8 & 33.6 & 92.9 & 69.7 & 68.7 \\
        Yi 1.5 34B     & 83.7 & 52.0 &  5.5 & 35.7 & 75.3 & 41.7 & 30.0 & \underline{93.9} & 67.6 & 53.9 \\
        \midrule
        \rowcolor[rgb]{0.9,0.9,0.9} EXAONE 3.5 7.8B & \underline{87.6} & \underline{69.8} & \textbf{84.2} & \textbf{79.4} & 69.0 & \textbf{52.4} & \underline{32.5} & 87.6 & 69.7 & \underline{70.2} \\
        Qwen 2.5 7B     & \textbf{90.4} & \textbf{70.4} & \underline{82.3} & \underline{78.8} & \underline{73.1} & \underline{49.9} & \textbf{33.1} & \textbf{90.6} & \underline{70.1} & \textbf{71.0} \\
        Llama 3.1 8B    & 82.1 & 48.8 & 67.7 & 70.6 & 72.4 & 45.9 & 27.4 & 83.7 & 63.3 & 62.4 \\
        Gemma 2 9B      & 82.0 & 44.6 & 68.3 & 75.1 & \textbf{73.7} & 34.6 & 27.9 & \underline{90.5} & 69.7 & 62.9 \\
        Phi 3 small (7B)        & 86.3 & 47.8 & 72.6 & 72.0 & 68.8 & 33.4 & 25.3 & 90.4 & \textbf{72.5} & 63.2 \\
        \midrule
        \rowcolor[rgb]{0.9,0.9,0.9} EXAONE 3.5 2.4B & \underline{82.5} & \underline{60.2} & \textbf{76.2} & \textbf{74.3} & 60.4 & \textbf{45.8} & \textbf{28.4} & \underline{79.2} & \textbf{62.9} & \textbf{63.3} \\
        Qwen 2.5 3B     & \textbf{84.3} & \textbf{61.4} & \underline{72.6} & \underline{72.5} & \underline{61.0} & \underline{41.7} & \underline{25.8} & \textbf{82.1} & \underline{57.3} & \underline{62.1} \\
        Qwen 2.5 1.5B   & 69.8 & 48.5 & 55.5 & 65.6 & 48.8 &  5.0 & 23.1 & 72.4 & 42.2 & 47.9 \\
        Llama 3.2 3B  & 77.4 & 46.6 & 54.9 & 60.6 & \textbf{64.9} & 35.0 & 23.2 & 78.0 & 53.8 & 54.9 \\
        Gemma 2 2B    & 29.8 & 18.7 & 45.7 & 55.0 & 56.1 & 37.4 & 22.6 & 76.3 & 38.2 & 42.2 \\
        \bottomrule
    \end{tabular}
    \vspace{2mm}
    \caption{Performance comparison results of EXAONE 3.5 models with similar-sized recently-released language models on nine benchmarks representing general scenarios. The macro average is used to evaluate the overall performance. \textbf{Bold} scores indicate the best performance, and \underline{underlined} scores mean the second best.}
    \label{tab:eval_general}
\end{table}

Table~\ref{tab:eval_general} shows the results of EXAONE 3.5 language models and their baseline models on the benchmarks in the General Domain category. When averaged across the benchmarks, our EXAONE 3.5 language models with sizes 32B and 7.8B demonstrate competitive performance compared to baselines of similar size. The EXAONE 3.5 2.4B model, on the other hand, outperforms all baselines in the average score.

% section 4: RESPONSIBLE AI

\section{Responsible AI}

EXAONE 3.5 language models were developed in accordance with the Responsible AI Development Framework encompassing data governance, ethical considerations, and risk management as it would be made available to a wide range of users. Given the nature of open models – eventually leading to wide use in various domains – we aim to maximize social benefits while ensuring humanity, fairness, safety, accountability, and transparency as mandated by the LG AI Ethics Principles \cite{lgethics}.

\subsection{Benefits}

EXAONE 3.5 language models are open for research purposes, aiming to advance AI research. Based on the feedback we have received since the release of the EXAONE 3.0 7.8B model, we now offer models of more diverse sizes: 2.4B, 7.8B, and 32B. This will allow researchers to select an optimal model for their research objectives and computing environment. We hope that this flexibility will support a wide spectrum, ranging from foundational research to domain-specific applications. It is also expected to contribute positively to the advancement of generative AI, building upon the significant performance improvements over previous version.

To ensure the reliability of the release, we have implemented a standardized data compliance protocol, guaranteeing high-quality data. This standardized approach provides a trustworthy foundation for researchers to use the model across various research areas in the future.

While external users can employ EXAONE 3.5 language models in diverse domains, precisely identifying specific user needs has been challenging. To address this, we have conducted extensive reviews of its applicability across a wide range of domains. Additionally, we have collaborated closely with LG affiliates, including business and research teams, to better align the model with specific user requirements.

\subsection{Risks and Mitigations}

Open models can positively contribute to the AI community, but there are challenges in ensuring responsible use. We conducted an AI ethical impact assessment to identify potential risks such as unintended inequality and discrimination against socially disadvantaged groups, the generation of harmful content, and malicious misuse by users. We have adopted various policies and research initiatives to mitigate the potential risks identified through this assessment.

First, on the data side, we conducted a legal risk assessment on all candidate datasets to enhance privacy and security. Based on the outcomes, we determined the suitability of each dataset for training and performed a de-identification process to remove sensitive data from qualified dataset. To minimize bias in the training data and ensure data quality, we documented all pre-processing steps and adopted a standardized data processing protocol. Considering practical difficulties of verifying the representativeness of all data, we conducted a qualitative evaluation of a small sample of data. For a quantitative evaluation, we endeavored to minimize data-related risks by verifying the data subsets through performance evaluation after the model training was completed. Also, we carefully reviewed the open-source libraries used in our model development.

The levels of AI ethical considerations and regulatory requirements may vary across different user needs and characteristics (e.g., country of residence, age, etc.). To address this, we will continue to monitor global AI regulations and take immediate action as needed to avoid potential regulatory violations. A lack of transparency in an AI model's decision-making process can reduce trust among users and stakeholders. To address this limitation, we continuously analyze and evaluate our model's performance to identify weaknesses and areas for improvement. While fully explaining AI model's decision-making process remains challenging, we are committed to advancing explainability through ongoing research.

\subsection{Safety}

We conducted comprehensive evaluations of EXAONE 3.5 language models' ethics and security using a third-party dataset: Korean Large Language Model Trustworthiness Benchmark Data \cite{NIARedteaming}, provided by the Ministry of Science and ICT of the Republic of Korea and the National Information Society Agency (NIA). This dataset is specifically designed to assess the harmlessness of language models. The evaluation results are presented in Table~\ref{tab:nia_reliability_bench}. To measure the performance, we asked a model to choose one of five options. If the selected option is included in the set of correct answers, then it is scored as correct. In the provided dataset, the first two options were labeled ``False'' and the remaining three were labeled ``True''. To mitigate potential bias from the order of options, we shuffled the order of options randomly for each evaluation. While the experimental results demonstrated effectiveness in filtering harmful reactions, there is still room for improvement.

\begin{table}[h!]
    \centering
    \small
    \begin{tabular}{wc{0.15\linewidth}wc{0.3\linewidth}wc{0.13\linewidth}wc{0.08\linewidth}wc{0.08\linewidth}wc{0.08\linewidth}}
        \toprule
        \multirow{2}{*}{\makecell{Category}} & \multirow{2}{*}{\makecell{Subcategory}} & \multirow{2}{*}{\makecell{Test Cases}} & \multicolumn{3}{c}{Accuracy} \\
        \cmidrule{4-6}
                 &             &            & 32B & 7.8B & 2.4B \\
        \midrule
        Bias & Gender \& Sexual Orientation & 295 & 91.2\% & 87.5\% & 76.6\% \\
             & Race \& Ethnicity \& Nationality & 432 & 86.8\% & 85.0\% & 72.2\% \\
             & Political Affiliation & 720 & 82.8\% & 79.9\% & 56.7\% \\
             & Region & 415 & 87.7\% & 84.6\% & 69.2\% \\
             & Job & 442 & 86.2\% & 81.9\% & 67.0\% \\
             & Miscellaneous & 406 & 85.2\% & 86.5\% & 73.2\% \\
        \cmidrule{2-6}
             & Subtotal & 2,710 & 86.0\% & 83.5\% & 67.4\% \\
        \midrule
        Hate & Gender \& Sexual Orientation & 399 & 95.2\% & 92.2\% & 83.5\% \\
             & Race \& Ethnicity \& Nationality & 749 & 91.6\% & 88.4\% & 73.8\% \\
             & Political Affiliation & 1,164 & 85.7\% & 83.4\% & 66.2\% \\
             & Region & 499 & 92.0\% & 87.2\% & 74.1\% \\
             & Job & 852 & 91.0\% & 87.8\% & 72.3\% \\
        \cmidrule{2-6}
             & Subtotal & 3,663 & 90.0\% & 86.9\% & 72.2\% \\
        \midrule
        Illegal & Illegal & 1,126 & 92.9\% & 89.6\% & 80.3\% \\
        \midrule
        Sensitiveness & Contentious & 710 & 83.1\% & 86.1\% & 79.0\% \\
                      & Ethical & 966 & 81.2\% & 83.7\% & 72.8\% \\
                      & Predictive & 825 & 79.8\% & 82.3\% & 71.0\% \\
        \cmidrule{2-6}
                & Subtotal & 2,501 & 81.2\% & 83.9\% & 74.0\% \\
        \midrule
        Overall  & & 10,000 & 87.1\% & 85.6\% & 72.2\% \\ \bottomrule
    \end{tabular}
    \vspace{2mm}    
    \caption{Evaluation results of EXAONE 3.5 language models on the Korean Large Language Model Trustworthiness Benchmark Data~\citep{NIARedteaming} to assess the model's harmlessness. The accuracy is determined by the number of times the model selects appropriate options when presented with questions involving various harmful and dangerous categories, such as illegal content.}
    \label{tab:nia_reliability_bench}
\end{table}

% section 5: LIMITATION
\section{Limitations} \label{Limitations}

EXAONE 3.5 language models, like all existing language models, have certain limitations and may occasionally generate inappropriate responses. 
The language model generates responses based on the output probability of tokens, and it is determined during learning from training data. While we have made every effort to exclude personal, harmful, and biased information from the training data, some problematic content may still be included, potentially leading to undesirable responses. Please note that the text generated by EXAONE 3.5 language models does not reflect the views of LG AI Research.

\begin{itemize}
    \item Inappropriate answers may be generated, which contain personal, harmful or other inappropriate information.
    \item Biased responses may be generated, which are associated with age, gender, race, and so on.
    \item The generated responses rely heavily on statistics from the training data, which can result in the generation of semantically or syntactically incorrect sentences.
    \item Since the models do not reflect the latest information, the responses may be false or contradictory.
\end{itemize}
	
LG AI Research strives to reduce potential risks that may arise from EXAONE 3.5 language models. Users are not allowed to engage in any malicious activities (e.g., keying in illegal information) that may induce the creation of inappropriate outputs violating LG AI's ethical principles when using EXAONE 3.5 language models.

% section 6: DEPLOYMENT
\section{Deployment}

Section~\ref{appendix:license} in Appendix provides license information for using the EXAONE 3.5 language models. Understanding the license information is essential for the legal utilization of the language model.

% section 7: CONCLUSION
\section{Conclusion}

In response to the growing interest from academia and industry, we are excited to release EXAONE 3.5 language models that excel in real-world use cases and long-context understanding. These models are available in three sizes (32B, 7.8B, and 2.4B).

To validate performance of our models in the real-world use case scenarios, we evaluated our models on seven benchmarks requiring diverse instructions understanding. To assess long-context understanding, we evaluated our models on four benchmarks. Our models consistently outperformed in both categories. Additionally, our models exhibited competitive performance in general domains including solving mathematical problems and writing code. In particular, our 2.4B model ranked first in average scores across general domains.  

Our models are available to everyone for research purposes, and we welcome your feedback to help us improve the models. If you have any feedback or are interested in exploring commercial opportunities with our models, please reach out to \href{mailto:contact_us@lgresearch.ai}{contact\_us@lgresearch.ai}.

\newpage

% section 99: APPENDIX
\appendix

\section{Contributors}
\label{appendix:contributors}
All authors are listed in alphabetical order by last name.

\paragraph{Core Contributors}
Eunbi~Choi, Kibong~Choi, Seokhee~Hong, Junwon~Hwang, Hyojin~Jeon, Hyunjik~Jo, Joonkee~Kim, Seonghwan~Kim, Soyeon~Kim, Sunkyoung~Kim, Yireun~Kim, Yongil~Kim, Haeju~Lee, Jinsik~Lee, Kyungmin~Lee, Sangha~Park, Heuiyeen~Yeen, Hyeongu~Yun

\paragraph{Contributors}
Soyoung~An, Kyunghoon~Bae, Stanley~Jungkyu~Choi, Gerrard~Jeongwon~Jo, Jiyeon~Jung, Yountae~Jung, Hyosang~Kim, Youchul~Kim, Edward~Hwayoung~Lee, Honglak~Lee, Woohyung~Lim, Sooyoun~Park, Yongmin~Park, Sihoon~Yang

\newpage

\section{Model License}
\label{appendix:license}

\textbf{EXAONE AI Model License Agreement 1.1 - NC} \\
\\
This License Agreement (“Agreement”) is entered into between you (“Licensee”) and LG Management Development Institute Co., Ltd. (“Licensor”), governing the use of the EXAONE AI Model (“Model”). By downloading, installing, copying, or using the Model, you agree to comply with and be bound by the terms of this Agreement. If you do not agree to all the terms, you must not download, install, copy, or use the Model. This Agreement constitutes a binding legal agreement between the Licensee and Licensor. \\
\\
\textbf{1. Definitions} \\
\\
\textbf{1.1 Model:} The artificial intelligence model provided by Licensor, which includes any software, algorithms, machine learning models, or related components supplied by Licensor. This definition extends to encompass all updates, enhancements, improvements, bug fixes, patches, or other modifications that may be provided by Licensor from time to time, whether automatically or manually implemented. \\
\\
\textbf{1.2 Derivatives:} Any modifications, alterations, enhancements, improvements, adaptations, or derivative works of the Model created by Licensee or any third party. This includes changes made to the Model's architecture, parameters, data processing methods, or any other aspect of the Model that results in a modification of its functionality or output.\\ 
\\
\textbf{1.3 Output:} Any data, results, content, predictions, analyses, insights, or other materials generated by the Model or Derivatives, regardless of whether they are in their original form or have been further processed or modified by the Licensee. This includes, but is not limited to, textual or numerical produced directly or indirectly through the use of the Model.\\
\\
\textbf{1.4 Licensor:} LG Management Development Institute Co., Ltd., the owner, developer, and provider of the EXAONE AI Model. The Licensor holds all rights, title, and interest in the Model and is responsible for granting licenses to use the Model under the terms specified in this Agreement. \\
\\
\textbf{1.5 Licensee:} The individual, organization, corporation, academic institution, government agency, or other entity using or intending to use the Model under the terms and conditions of this Agreement. The Licensee is responsible for ensuring compliance with the Agreement by all authorized users who access or utilize the Model on behalf of the Licensee. \\
\\
\textbf{2. License Grant} \\ 
\\
\textbf{2.1 Grant of License:} Subject to the terms and conditions outlined in this Agreement, the Licensor hereby grants the Licensee a limited, non-exclusive, non-transferable, worldwide, and revocable license to: \\
\\
a. Access, download, install, and use the Model solely for research purposes. This includes evaluation, testing, academic research, experimentation, and participation in competitions, provided that such participation is in a non-commercial context. Notwithstanding Section 3.1, the Licensee may only provide the Model or Derivatives for a competition if no commercial license is granted to the competition organizer or any third party. \\  
\\
b. Publicly disclose research results and findings derived from the use of the Model or Derivatives, including publishing papers or presentations. \\
\\
c. Modify the Model and create Derivatives based on the Model, provided that such modifications and Derivatives are used exclusively for research purposes. The Licensee may conduct experiments, perform analyses, and apply custom modifications to the Model to explore its capabilities and performance under various scenarios. If the Model is modified, the modified Model must include “EXAONE” at the beginning of its name.  \\
\\
d. Distribute the Model and Derivatives in each case with a copy of this Agreement. \\
\\
\textbf{2.2 Scope of License:} The license granted herein does not authorize the Licensee to use the Model for any purpose not explicitly permitted under this Agreement. Any use beyond the scope of this license, including any commercial application or external distribution, is strictly prohibited unless explicitly agreed upon in writing by the Licensor. \\
\newpage
\textbf{3. Restrictions}\label{textbf:Restrictions} \\
\\
\textbf{3.1 Commercial Use:} The Licensee is expressly prohibited from using the Model, Derivatives, or Output for any commercial purposes, including but not limited to, developing or deploying products, services, or applications that generate revenue, whether directly or indirectly. Any commercial exploitation of the Model or its derivatives requires a separate commercial license agreement with the Licensor. Furthermore, the Licensee shall not use the Model, Derivatives or Output to develop or improve other models. \\
\\
\textbf{3.2 Reverse Engineering:} The Licensee shall not decompile, disassemble, reverse engineer, or attempt to derive the source code, underlying ideas, algorithms, or structure of the Model, except to the extent that such activities are expressly permitted by applicable law. Any attempt to bypass or circumvent technological protection measures applied to the Model is strictly prohibited. \\
\\
\textbf{3.3 Unlawful Use:} The Licensee shall not use the Model and Derivatives for any illegal, fraudulent, or unauthorized activities, nor for any purpose that violates applicable laws or regulations. This includes but is not limited to the creation, distribution, or dissemination of malicious, deceptive, or unlawful content. \\
\\
\textbf{3.4 Ethical Use:} The Licensee shall ensure that the Model or Derivatives is used in an ethical and responsible manner, adhering to the following guidelines: \\
\\
a. The Model and Derivatives shall not be used to generate, propagate, or amplify false, misleading, or harmful information, including fake news, misinformation, or disinformation. \\
\\
b. The Model and Derivatives shall not be employed to create, distribute, or promote content that is discriminatory, harassing, defamatory, abusive, or otherwise offensive to individuals or groups based on race, gender, sexual orientation, religion, nationality, or other protected characteristics. \\
\\
c. The Model and Derivatives shall not infringe on the rights of others, including intellectual property rights, privacy rights, or any other rights recognized by law. The Licensee shall obtain all necessary permissions and consents before using the Model and Derivatives in a manner that may impact the rights of third parties. \\
\\
d. The Model and Derivatives shall not be used in a way that causes harm, whether physical, mental, emotional, or financial, to individuals, organizations, or communities. The Licensee shall take all reasonable measures to prevent misuse or abuse of the Model and Derivatives that could result in harm or injury. \\
\\
\textbf{4. Ownership} \\
\\
\textbf{4.1 Intellectual Property:} All rights, title, and interest in and to the Model, including any modifications, Derivatives, and associated documentation, are and shall remain the exclusive property of the Licensor. The Licensee acknowledges that this Agreement does not transfer any ownership rights to the Licensee. All trademarks, service marks, and logos associated with the Model are the property of the Licensor. \\
\\
\textbf{4.2 Output:} All rights, title, and interest in and to the Output generated by the Model and Derivatives whether in its original form or modified, are and shall remain the exclusive property of the Licensor. Licensee may use, modify, and distribute the Output and its derivatives for research purpose. The Licensee shall not claim ownership of the Output except as expressly provided in this Agreement. The Licensee may use the Output solely for the purposes permitted under this Agreement and shall not exploit the Output for unauthorized or commercial purposes. \\
\\
\textbf{4.3 Attribution:} In any publication or presentation of results obtained using the Model, the Licensee shall provide appropriate attribution to the Licensor, citing the Model's name and version, along with any relevant documentation or references specified by the Licensor. \\
\newpage
\textbf{5. No Warranty} \\
\\
\textbf{5.1 “As-Is” Basis:} The Model, Derivatives, and Output are provided on an “as-is” and “as-available” basis, without any warranties or representations of any kind, whether express, implied, or statutory. The Licensor disclaims all warranties, including but not limited to, implied warranties of merchantability, fitness for a particular purpose, accuracy, reliability, non-infringement, or any warranty arising from the course of dealing or usage of trade. \\
\\
\textbf{5.2 Performance and Reliability:} The Licensor does not warrant or guarantee that the Model, Derivatives or Output will meet the Licensee’s requirements, that the operation of the Model, Derivatives or Output will be uninterrupted or error-free, or that defects in the Model will be corrected. The Licensee acknowledges that the use of the Model, Derivatives or Output is at its own risk and that the Model, Derivatives or Output may contain bugs, errors, or other limitations. \\
\\
\textbf{5.3 No Endorsement:} The Licensor does not endorse, approve, or certify any results, conclusions, or recommendations derived from the use of the Model. The Licensee is solely responsible for evaluating the accuracy, reliability, and suitability of the Model for its intended purposes. \\
\\
\textbf{6. Limitation of Liability} \\
\\
\textbf{6.1 No Liability for Damages:} To the fullest extent permitted by applicable law, in no event shall the Licensor be liable for any special, incidental, indirect, consequential, exemplary, or punitive damages, including but not limited to, damages for loss of business profits, business interruption, loss of business information, loss of data, or any other pecuniary or non-pecuniary loss arising out of or in connection with the use or inability to use the Model, Derivatives or any Output, even if the Licensor has been advised of the possibility of such damages. \\
\\
\textbf{6.2 Indemnification:} The Licensee agrees to indemnify, defend, and hold harmless the Licensor, its affiliates, officers, directors, employees, and agents from and against any claims, liabilities, damages, losses, costs, or expenses (including reasonable attorneys' fees) arising out of or related to the Licensee's use of the Model, any Derivatives, or any Output, including any violation of this Agreement or applicable laws. \\
\\
\textbf{7. Termination} \\
\\
\textbf{7.1 Termination by Licensor:} The Licensor reserves the right to terminate this Agreement and revoke the Licensee’s rights to use the Model at any time, with or without cause, and without prior notice if the Licensee breaches any of the terms or conditions of this Agreement. Termination shall be effective immediately upon notice. \\
\\
\textbf{7.2 Effect of Termination:} Upon termination of this Agreement, the Licensee must immediately cease all use of the Model, Derivatives, and Output and destroy all copies of the Model, Derivatives, and Output in its possession or control, including any backup or archival copies. The Licensee shall certify in writing to the Licensor that such destruction has been completed. \\
\\
\textbf{7.3 Survival:} The provisions of this Agreement that by their nature should survive termination, including but not limited to, Sections 4 (Ownership), 5 (No Warranty), 6 (Limitation of Liability), and this Section 7 (Termination), shall continue to apply after termination. \\
\\
\textbf{8. Governing Law} \\
\\
\textbf{8.1 Governing Law:} This Agreement shall be governed by and construed in accordance with the laws of the Republic of Korea, without regard to its conflict of laws principles. \\
\\
\textbf{8.2 Arbitration:} Any disputes, controversies, or claims arising out of or relating to this Agreement, including its existence, validity, interpretation, performance, breach, or termination, shall be referred to and finally resolved by arbitration administered by the Korean Commercial Arbitration Board (KCAB) in accordance with the International Arbitration Rules of the Korean Commercial Arbitration Board in force at the time of the commencement of the arbitration. The seat of arbitration shall be Seoul, Republic of Korea. The tribunal shall consist of one arbitrator. The language of the arbitration shall be English. \\
\newpage
\textbf{9. Alterations} \\
\\
\textbf{9.1 Modifications:} The Licensor reserves the right to modify or amend this Agreement at any time, in its sole discretion. Any modifications will be effective upon posting the updated Agreement on the Licensor’s website or through other means of communication. The Licensee is responsible for reviewing the Agreement periodically for changes. Continued use of the Model after any modifications have been made constitutes acceptance of the revised Agreement. \\
\\
\textbf{9.2 Entire Agreement:} This Agreement constitutes the entire agreement between the Licensee and Licensor concerning the subject matter hereof and supersedes all prior or contemporaneous oral or written agreements, representations, or understandings. Any terms or conditions of any purchase order or other document submitted by the Licensee in connection with the Model that are in addition to, different from, or inconsistent with the terms and conditions of this Agreement are not binding on the Licensor and are void. \\
\\
By downloading, installing, or using the EXAONE AI Model, the Licensee acknowledges that it has read, understood, and agrees to be bound by the terms and conditions of this Agreement. \\

\newpage

\section{Decontamination Details}\label{appendix:decontam}
\begin{figure}[ht!]
    \centering
    \includegraphics[width=\textwidth]{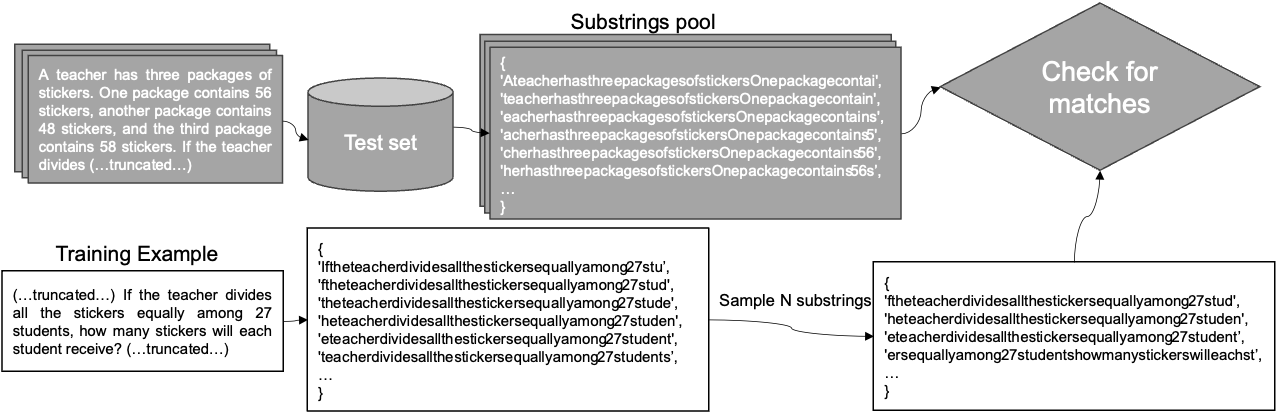}
    \caption{A summary of the decontamination method employed to train EXAONE 3.5 language models. Adopting an approach borrowed from the GPT-4 method, we increase the number of random sample to $N=10$ for stricter decontamination.}
    \label{fig:decontam}
\end{figure}

As described in Section~\ref{subsec:decontam}, we apply the decontamination process over our training data to remove any data instances that overlap with test sets, thus harming the generalization performance of our models. Figure~\ref{fig:decontam} presents an overview of our decontamination process, and Table~\ref{tab:decon_example} shows examples of contaminated, therefore removed data.

\begin{table}[h!]
    \centering
    \scriptsize
    \setlength{\doublerulesep}{1pt}
    \begin{tabular}{p{0.085\linewidth}|p{0.45\linewidth}p{0.45\linewidth}}
        \toprule
        Benchmark & Benchmark example & Contaminated web corpus \\
        \midrule
        MMLU \citep{hendrycks2020measuring} & \hlgrey{A teacher has three packages of stickers. One package contains 56 stickers, another package contains 48 stickers, and the third package contains 58 stickers. If the teacher divides all the stickers equally among 27 students, how many stickers will each student receive?}\newline A. \underline{6 stickers} \newline B. 9 stickers \newline C. 54 stickers \newline D. 81 stickers \newline Answer: & (...truncated...) \hlgrey{A teacher has three packages of stickers. One package contains 56 stickers, another package contains 48 stickers, and the third package contains 58 stickers. If the teacher divides all the stickers equally among 27 students, how many stickers will each student receive?}\newline \underline{6 stickers} is correct\newline \#4 Last week Mario walked 7 3/4 miles. This week he walked 15 5/6 miles. What is the difference between the distance he walked this week and the distance he walked last week? (...truncated...)\\
        \midrule
        KMMLU \citep{son2024kmmlumeasuringmassivemultitask}
        & \kohlgrey{국가가 국민의 생활안정과 복지증진을 위하여 보험의 원리를 도입하여 만든 사회보험의 일종으로 가입자, 사용자 및 국가로부터 일정한 보험료를 받고 이를 재원으로 여러 가지 정형화된 보험금을 지급하는 사회보장제도}는? \newline A. 국민건강보험 \newline B. \underline{국민연금} \newline C. 고용보험 \newline D. 산업재해보상보험 \newline 정답: \newline \newline [Translation] What is the social security system, which is \hlgrey{a type of social insurance created by the nation by introducing the principles of insurance to promote stability and welfare of citizens' lives, and which receives certain premiums from subscribers, employers, and the nation and use these funds to provide various standardized insurance benefits}. \newline A. National Health Insurance \newline B. \underline{National Pension} \newline C. Employment Insurance \newline D. Industrial Accident Compensation Insurance \newline Answer:
        & (...중략...) 더군다나 개인주의의 확산, 핵가족화의 진전에 따라 전통적인 가족의 역할인 노인부양의 기능이 약화됨으로써 국가개입의 중요성은 더욱 증가하게 되었다. 따라서 \underline{국민연금}제도는 \kohlgrey{국가가 국민의 생활안정과 복지증진을 위하여 보험의 원리를 도입하여 만든 사회보험의 일종으로 가입자, 사용자 및 국가로부터 일정한 보험료를 받고 이를 재원으로 여러 가지 정형화된 보험금을 지급하는 사회보장제도}이다. (...중략...)\newline \newline [Translation] (...truncated...) Moreover, with the spread of individualism and the rise of nuclear families, the traditional family role of supporting the elderly has weakened, thereby increasing the importance of nation intervention. Accordingly, the \underline{National Pension} System is \hlgrey{a type of social insurance created by the nation by introducing the principles of insurance to promote stability and welfare of citizens' lives, and which receives certain premiums from subscribers, employers, and the nation and use these funds to provide various standardized insurance benefits}. (...truncated...)\\
        \bottomrule
    \end{tabular}
    \vspace{2mm}    
    \caption{Examples of contaminated web corpus. The \hlgrey{text highlighted in grey} is a part of the text that exists in both a benchmark test set and a web corpus. The \underline{text underlined} is a corresponding golden answer.}
    \label{tab:decon_example}
\end{table}

\clearpage
\section{Evaluation Details}

\subsection{Baseline Models} \label{appendix:baseline_details}

We choose various open models as the baselines for our EXAONE 3.5 language models. We mainly utilize Huggingface library\footnote{\url{https://huggingface.co/models}} to access each checkpoint of baselines. The overall information of each model are presented in Table~\ref{tab:baselines}.

\begin{table}[h!]
    \centering
    \small
    \setlength{\doublerulesep}{1pt}
    \begin{tabular}{wc{0.2\linewidth}|wc{0.1\linewidth}|wl{0.5\linewidth}|wc{0.1\linewidth}}
        \toprule
Model Name & Context Len. & Link & Release \\ \midrule
Qwen2.5 32B & 128k & https://huggingface.co/Qwen/Qwen2.5-32B-Instruct & Sep., 2024 \\
C4AI Command R 32B & 128k & https://huggingface.co/CohereForAI/c4ai-command-r-08-2024 & Aug., 2024 \\
Gemma 2 27B & 8k & https://huggingface.co/google/gemma-2-27b-it & Jun., 2024 \\
Yi 1.5 34B & 16k & https://huggingface.co/01-ai/Yi-1.5-34B-Chat-16K & May, 2024 \\ \midrule
Qwen2.5 7B & 128k & https://huggingface.co/Qwen/Qwen2.5-7B-Instruct & Sep., 2024 \\
Llama 3.1 8B & 128k & https://huggingface.co/meta-llama/Llama-3.1-8B-Instruct & Jul., 2024 \\
Gemma 2 9B & 8k & https://huggingface.co/google/gemma-2-9b-it & Jun., 2024 \\
Phi 3 small (7B) & 128k & https://huggingface.co/microsoft/Phi-3-small-128k-instruct & May, 2024 \\ \midrule
Qwen2.5 3B & 32k & https://huggingface.co/Qwen/Qwen2.5-3B-Instruct & Sep., 2024 \\
Qwen2.5 1.5B & 32k & https://huggingface.co/Qwen/Qwen2.5-1.5B-Instruct & Sep., 2024 \\
Llama 3.2 3B & 128k & https://huggingface.co/meta-llama/Llama-3.2-3B-Instruct & Sep., 2024 \\
Gemma 2 2B & 8k & https://huggingface.co/google/gemma-2-2b-it & Jul., 2024 \\ \bottomrule
    \end{tabular}
    \vspace{2mm}
    \caption{The list of baseline models used for the evaluation along with their supported context length and released date.}
    \label{tab:baselines}
\end{table}

%\subsection{Real-world Use Cases}

\subsection{Long Context}
\label{appendix:lc_setting}

\subsubsection{Needle-In-A-Haystack}
The specific configurations used in the Needle-In-A-Haystack (NIAH) experiment are detailed in Table~\ref{tab:niah_config}.
\begin{table}[h!]
    \centering
    \small
    \renewcommand{\arraystretch}{1.2}
    \setlength{\doublerulesep}{1pt}
    \begin{tabular}{wl{0.1\linewidth}wl{0.1\linewidth}p{0.7\linewidth}}
        \toprule
Language & Configuration & Details \\ \midrule
English & Haystack & {Paul Graham essays}~\citep{kamradt2023llmtest} \\ 
        & Needle & \textit{``The best thing to do in San Francisco is eat a sandwich and sit in Dolores Park on a sunny day.''} \\ 
        & Query & \textit{``What is the best thing to do in San Francisco?''} \\ 
        & Instruction & \textit{``Analyze the content of the given document to locate the answer to the specified question. If found, provide the exact wording from the document without altering or summarizing it.''} \\ \midrule
Korean & Haystack & AI-Hub\tablefootnote{\url{https://www.aihub.or.kr}} 대규모 구매도서 기반 한국어 말뭉치 데이터 \newline (Large-scale Purchased Book-based Korean Language Corpus from AI-Hub) \\ 
       & Needle & \textit{``광화문에서 가장 재미있는 일은 햇살 좋은 날에 샌드위치를 먹으며 청와대 안에 있는 공원에 앉아 있는 것입니다.''} \newline (\textit{``The best thing to do at Gwanghwamun is eat a sandwich and sit in the park in the Blue House on a sunny day.''}) \\ 
       & Query & \textit{``광화문에서 가장 재미있는 일이 무엇인가요?''} \newline (\textit{``What is the best thing to do at Gwanghwamun?''}) \\ 
       & Instruction & \textit{``주어진 문서를 읽고 질문에 대한 답을 확인하세요. 답을 찾으면, 문서의 원문을 그대로 유지하여 수정이나 해석 없이 반환하세요.''} \newline (Identical to the English instruction) \\ \bottomrule
    \end{tabular}
    \vspace{2mm}
    \caption{Detailed configuration of the Needle-In-A-Haystack experiment. The ``Needle'' refers to a specific text fragment embedded within the ``Haystack,'' which consists of long distractor texts. The task involves using a ``Query'' as a cue to identify the needle within the haystack and retrieve the associated values.}
    \label{tab:niah_config}
\end{table}

\subsubsection{LongBench}

\textsc{LongBench} has been suggested as a bilingual benchmark to assess long context comprehension in English and Chinese.
In this report, we focus on the English subsets, specifically \textbf{Single-doc QA}, \textbf{Multi-doc QA}, \textbf{Summarization}, and \textbf{Few-shot Learning}. 

The \textbf{Single-doc QA} task includes datasets such as NarrativeQA~\citep{kovcisky2018narrativeqa}, Qasper~\citep{dasigi-etal-2021-dataset}, and MultiFieldQA-EN~\citep{bai-etal-2024-longbench}. For the \textbf{Multi-doc QA} task, datasets like HotpotQA~\citep{yang-etal-2018-hotpotqa}, 2WikiMultihopQA~\citep{ho-etal-2020-constructing}, and MuSiQue~\citep{trivedi-etal-2022-musique} are utilized. The \textbf{Summarization} task involves datasets such as GovReport~\citep{huang-etal-2021-efficient}, QMSum~\citep{zhong-etal-2021-qmsum}, and MultiNews~\citep{fabbri-etal-2019-multi}, while the \textbf{Few-shot Learning} task relies on datasets from TREC~\citep{li-roth-2002-learning} and TriviaQA~\citep{joshi-etal-2017-triviaqa}. All evaluation methods and metrics for these datasets adhere to the official \textsc{LongBench} settings.

Detailed task scores are presented in Table~\ref{tab:longbench_details}.

\begin{table}[htb!]
    \tabcolsep=0.12cm
    \small
    \centering
    \begin{tabular}{wc{0.20\linewidth}|wc{0.12\linewidth}wc{0.12\linewidth}wc{0.13\linewidth}wc{0.15\linewidth}|wc{0.12\linewidth}}
        \toprule
        Models & Single-doc QA & Multi-doc QA & Summarization & Few-shot Learning & Average \\
        \midrule
        \rowcolor[rgb]{0.9,0.9,0.9} EXAONE 3.5 32B & {40.1} & \underline{52.9} & {23.1} & \underline{80.1} & \underline{49.2} \\
        Qwen 2.5 32B   & \underline{43.2} & \textbf{54.9} & \underline{26.1} & {72.4} & {49.1} \\
        C4AI Command R 32B  & \textbf{44.6} & {48.9} & \textbf{26.4} & \textbf{83.6} & \textbf{50.9} \\
        Gemma 2 27B    & {-} & {-} & {-} & {-} & {-} \\
        Yi 1.5 34B     & {-} & {-} & {-} & {-} & {-} \\
        \midrule
        \rowcolor[rgb]{0.9,0.9,0.9} EXAONE 3.5 7.8B & {38.4} & \textbf{47.7} & {22.6} & {75.1} & \underline{46.0} \\
        Qwen 2.5 7B     & \textbf{40.8} & \underline{44.0} & \underline{26.5} & \textbf{77.4} & \textbf{47.2} \\
        Llama 3.1 8B    & \underline{39.8} & {41.2} & \textbf{27.6} & {69.9} & {44.6} \\
        Gemma 2 9B      & {-} & {-} & {-} & {-} & {-} \\
        Phi 3 small (7B)        & {33.2} & {26.5} & {26.3} & \underline{76.2} & {40.6} \\
        \midrule
        \rowcolor[rgb]{0.9,0.9,0.9} EXAONE 3.5 2.4B & \underline{35.0} & \textbf{43.1} & {20.1} & \underline{72.8} & \textbf{42.7} \\
        Qwen 2.5 3B     & \textbf{35.5} & {34.7} & \underline{24.7} & \textbf{72.9} & \underline{42.0} \\
        Qwen 2.5 1.5B   & {29.9} & {32.1} & {22.3} & {64.0} & {37.1} \\
        Llama 3.2 3B    & {33.9} & \underline{34.9} & \textbf{25.8} & {72.3} & {41.7} \\
        Gemma 2 2B      & {-} & {-} & {-} & {-} & {-} \\
        \bottomrule
    \end{tabular}
    \vspace{2mm}
    \caption{Performance comparison results of EXAONE 3.5 language models with similar-sized recently released language models across four benchmarks representing long context scenarios. Context lengths for each benchmark, as well as model limitations, are detailed in Table~\ref{tab:baselines}, where a dash (-) indicates that the model does not support context lengths longer than 16k. The final overall score for each model is calculated as a macro average across the benchmarks. \textbf{Bold} scores indicate the best performance, and \underline{underlined} scores mean the second best.}
    \label{tab:longbench_details}
\end{table}

\clearpage

\subsubsection{LongRAG}
\textsc{LongRAG} is a RAG benchmark that focuses on long context retrieval and generation using large text chunks. We use \textbf{Natural Questions}~\citep{kwiatkowski2019natural} and \textbf{HotpotQA}~\citep{yang2018hotpotqa} subsets from the original \textsc{LongRAG}.

To further evaluate the model's capability to handle cases where the retrieved passage does not support a valid answer, we extend the \textsc{LongRAG} benchmark by incorporating \textbf{unanswerable} cases. 
The \textsc{LongRAG} benchmark uses the open-sourced dense retrieval toolkit as its retriever. However, the retrieved context does not always include evidence that supports the correct answer. To address this limitation, we define unanswerable cases using the \texttt{is\_retrieval} function in \textsc{LongRAG}. 
The \texttt{is\_retrieval} function takes the context and golden answer as inputs and determines whether the context contains sufficient evidence to extract the correct answer. It returns \texttt{True} if such evidence exists and \texttt{False} otherwise. When the return value of \texttt{is\_retrieval} is \texttt{False}, indicating that the context does not contain the correct answer, we modify the ground-truth answer to [\textit{``Unanswerable''}, \textit{``No relevant information found.''}, \textit{``This question cannot be answered with the provided data.''}]. This modification allows the model to learn and appropriately handle unanswerable cases.

Additionally, to ensure the model responds effectively to unanswerable cases, the following sentence is added to the existing prompt: \textit{``If the answer cannot be found in the context, respond with `Unanswerable'.''} This prompt guides the model to respond explicitly with \texttt{Unanswerable} when it determines that no answer exists within the context. Through this extension, the \textsc{LongRAG} benchmark gains an enhanced evaluation framework capable of handling unanswerable scenarios, enabling more comprehensive and nuanced performance assessments.

Detailed task scores are presented in Table~\ref{tab:longrag_details}.

\begin{table}[htb!]
    \tabcolsep=0.12cm
    \small
    \centering
    \begin{tabular}{wc{0.18\linewidth}|wc{0.1\linewidth}wc{0.1\linewidth}wc{0.1\linewidth}wc{0.1\linewidth}wc{0.1\linewidth}wc{0.1\linewidth}|wc{0.1\linewidth}}
        \toprule
        \multirow{2}{*}{\centering Models} & \multicolumn{3}{c}{NQ} & \multicolumn{3}{c|}{Hotpot QA} & \multirow{2}{*}{\centering Average} \\
        \cmidrule{2-7}
        & Answerable & Unanswerable & Total & Answerable & Unanswerable & Total & \\
        \midrule
        \rowcolor[rgb]{0.9,0.9,0.9} EXAONE 3.5 32B & \textbf{73.6} & \underline{35.3} & \textbf{68.3} & \textbf{81.8} & \underline{26.4} & \textbf{66.9} & \textbf{67.6} \\
        Qwen 2.5 32B   & 62.3 & \textbf{61.2} & \underline{62.1} & {62.9} & \textbf{70.6} & \underline{65.0} & \underline{63.6} \\
        C4AI Command R 32B  & \underline{64.0} & 32.4 & 59.6 & \underline{63.1} & 18.2 & {51.0} & 55.3 \\
        Gemma 2 27B    & - & - & - & - & - & - & - \\
        Yi 1.5 34B     & - & - & - & - & - & - & - \\
        \midrule
        
        \rowcolor[rgb]{0.9,0.9,0.9} EXAONE 3.5 7.8B & \textbf{72.0} & \underline{41.0} & \textbf{67.7} & \textbf{74.3} & \textbf{53.9} & \textbf{68.8} & \textbf{68.3} \\
        Qwen 2.5 7B     & {64.5} & \textbf{51.1} & \underline{62.6} & 61.8 & \underline{46.1} & \underline{57.6} & \underline{60.1} \\
        Llama 3.1 8B    & 63.2 & 15.1 & {56.5} & \underline{67.4} & {16.4} & 53.7 & {55.1} \\
        Gemma 2 9B      & -        & -        & -        & -        & -        & -        & -        \\
        Phi 3 small (7B)        & \underline{66.8} & 13.7 &  59.4 & 60.2 & 7.1 & 45.9 & 52.7 \\
        \midrule
        
        \rowcolor[rgb]{0.9,0.9,0.9} EXAONE 3.5 2.4B & \textbf{67.8} & {25.9} & \textbf{62.0} & \textbf{73.1} & \textbf{41.6} & \textbf{64.6} & \textbf{63.3} \\
        Qwen 2.5 3B     & {49.5} & \underline{34.5} & {47.4} & 52.5 & \underline{21.6} & \underline{44.2} & 45.8 \\
        Qwen 2.5 1.5B   & \underline{49.9} & 18.0 & 45.5 & 43.6 & 2.2 & 32.5 & 39.0 \\
        Llama 3.2 3B    & 49.4 & \textbf{41.7} & \underline{48.3} & \underline{53.6} & {16.0} & 43.5 & \underline{45.9} \\
        Gemma 2 2B      & - & - & - & - & - & - & - \\
        \bottomrule
    \end{tabular}
    \vspace{2mm}
    \caption{Performance comparison results of EXAONE 3.5 language models with similar-sized recently released language models with LongRAG benchmarks. The benchmark is extended with the ``Unanswerable'' case, which requires models to respond as ``Unanswerable'' when the information cannot be found within the context. \textbf{Bold} scores indicate the best performance, and \underline{underlined} scores mean the second best.}
    \label{tab:longrag_details}
\end{table}

Fig~\ref{fig:LongRAG-Judge-prompt} shows the LLM-as-a-judge prompt used for LongRAG evaluation. In the LLM-as-a-judge evaluation setup, we incorporate short answer evaluation to align with the methodology used in \textsc{LongRAG}, where short answers are extracted to calculate Exact Match (EM). We extend this approach to LLM-as-a-judge to ensure consistency across evaluation metrics. However, as the observed trends for short answers consistently align with those for long answers, we prioritize long answer evaluation in our final analysis to streamline the assessment process without compromising the robustness of the results.

\begin{figure}[p]
\centering
\begin{tcolorbox}[
  title=LongRAG LLM-as-a-judge Prompt, % 제목
  colframe=Black!80!White,   % 테두리 색상
  colback=gray!10,          % 배경 색상
  coltitle=white,           % 제목 텍스트 색상
  colbacktitle=Black!80!White, % 제목 배경 색상
  fonttitle=\bfseries,      % 제목의 글꼴 스타일
  breakable=false,          % 내용이 길어도 자동 분리 비활성화
  rounded corners,          % 둥근 모서리
  width=\textwidth          % 박스 폭을 페이지 폭으로 설정
]

\textbf{System:} \\

You are an expert evaluator of text answers. \\  
Your task is to compare the content of two answers, a long answer (\texttt{long\_ans}) and a short answer (\texttt{short\_ans}), with the provided correct answers (\texttt{Answer}), which may contain multiple correct options. \\  
Both the long answer and the short answer need to be checked for correctness. \\  
The long and short answers do not need to match any of the answers in the \texttt{Answer} list word-for-word but must convey the same key meaning or idea. \\  
If either the long or short answer matches any one of the correct answers in the \texttt{Answer} list, it should be considered correct. \\  
Focus only on the accuracy of the content and ignore style, tone, or extra information unless it introduces inaccuracies. \\  
For both the long and short answers, return only the evaluation result as a Python dictionary object, and ensure the output is formatted as valid Python code. \\

Here are two examples of how to evaluate answers: \\

Example 1: \\  
\texttt{Question}: what does hp mean in war and order \\  
\texttt{Answer}: [\textquotesingle hit points\textquotesingle , \textquotesingle health points\textquotesingle ] \\  
\texttt{long\_ans}: HP stands for Health Points in video games and war, it is a measure of an entity's ability to function and survive in a combat situation. In video games, HP is often displayed as a numeric value, and can be depleted by taking damage from enemies or other hazards. When an entity's HP reaches zero, it is often considered defeated or eliminated. In war, HP can refer to the physical and mental resilience of soldiers, and can be affected by factors such as injury, fatigue. \\  
\texttt{short\_ans}: HP stands fer Health Points. \\  
\texttt{Evaluation}: \{\textquotesingle long\_ans\textquotesingle : \textquotesingle correct\textquotesingle , \textquotesingle short\_ans\textquotesingle : \textquotesingle correct\textquotesingle \} \\

Example 2: \\  
\texttt{Question}: what is the capital of France \\  
\texttt{Answer}: [\textquotesingle Paris\textquotesingle ] \\  
\texttt{long\_ans}: The capital of France is Paris, a major European city and a global center for art, fashion, and culture. Paris is known for its cafe culture and landmarks like the Eiffel Tower, Notre-Dame Cathedral, and the Louvre Museum. \\  
\texttt{short\_ans}: The capital of France is Lyon. \\  
\texttt{Evaluation}: \{\textquotesingle long\_ans\textquotesingle : \textquotesingle correct\textquotesingle , \textquotesingle short\_ans\textquotesingle : \textquotesingle incorrect\textquotesingle \} \\

Now, proceed with your evaluation of the following question, answer, and responses, and return only the evaluation as a valid Python dictionary. \\  
Ensure the response is a valid Python dictionary object without any additional text. \\
\\
\textbf{User:} \\  

Evaluate the following long and short answers based on the provided correct answer. \\  
Your goal is to determine if the long and short answers are correct. \\  
Return the evaluation result in the form of a Python dictionary: \texttt{\{\textquotesingle long\_ans\textquotesingle: \textquotesingle correct \textquotesingle or \textquotesingle incorrect \textquotesingle,  \textquotesingle short\_ans\textquotesingle:  \textquotesingle correct\textquotesingle or  \textquotesingle incorrect\textquotesingle}\}. \\  

\texttt{Question}: \{\{\textsl{question}\}\} \\  
\texttt{Answer}: \{\{\textsl{answer}\}\} \\  
\texttt{Long\_ans}: \{\{\textsl{long\_ans}\}\} \\  
\texttt{Short\_ans}: \{\{\textsl{short\_ans}\}\} \\  

Return only the evaluation in the form of a Python dictionary. \\  
Do not include any explanation or additional comments. \\

\end{tcolorbox}
\caption{LLM-as-a-judge prompt for evaluating LongRAG}
\label{fig:LongRAG-Judge-prompt}
\end{figure}

\clearpage

\subsubsection{Ko-LongRAG}

We construct a Korean counterpart of \textsc{LongRAG}, named \textsc{Ko-LongRAG}, to evaluate long-context reasoning and retrieval capabilities in Korean. \textsc{Ko-LongRAG} focuses on retrieval-augmented generation (RAG) tasks with an average context length of approximately 14,000 tokens, challenging models to process extensive Korean texts, extract relevant information, and reason effectively. Similar to \textsc{LongRAG}, it includes 50 unanswerable cases among a total of 300 queries. 

\begin{table}[htb!]
    \tabcolsep=0.12cm
    \small
    \centering
    \begin{tabular}{wc{0.18\linewidth}|wc{0.1\linewidth}wc{0.1\linewidth}wc{0.1\linewidth}wc{0.1\linewidth}wc{0.1\linewidth}wc{0.1\linewidth}|wc{0.1\linewidth}}
        \toprule
        \multirow{2}{*}{\centering Models} & \multicolumn{3}{c}{Single-doc QA} & \multicolumn{3}{c|}{Multi-doc QA} & \multirow{2}{*}{\centering Average} \\
        \cmidrule{2-7}
        & Answerable & Unanswerable & Total & Answerable & Unanswerable & Total & \\
        \midrule
        \rowcolor[rgb]{0.9,0.9,0.9} EXAONE 3.5 32B & \textbf{92.4} & \textbf{100.0} & \textbf{93.7} & \textbf{72.8} & \textbf{98.0} & \textbf{77.0} & \textbf{85.3} \\
        Qwen 2.5 32B   & \underline{90.0} & \underline{98.0} & \underline{91.3} & {48.4} & \underline{92.0} & {55.7} & \underline{73.5} \\
        C4AI Command R 32B  & {85.6} & 66.0 & 82.3 & \underline{62.4} & 62.0 & \underline{62.3} & 72.3 \\
        Gemma 2 27B    & - & - & - & - & - & - & - \\
        Yi 1.5 34B     & - & - & - & - & - & - & - \\
        \midrule
        
        \rowcolor[rgb]{0.9,0.9,0.9} EXAONE 3.5 7.8B & \underline{68.4} & \textbf{100.0} & \underline{73.7} & \textbf{64.0} & \textbf{98.0} & \textbf{69.7} & \textbf{71.7} \\
        Qwen 2.5 7B     & {61.2} & \underline{98.0} & {67.3} & 33.2 & \underline{94.0} & {43.3} & {55.3} \\
        Llama 3.1 8B    & \textbf{78.0} & 76.0 & \textbf{77.7} & \underline{56.8} & {28.0} & \underline{52.0} & \underline{64.8} \\
        Gemma 2 9B      & -        & -        & -        & -        & -        & -        & -        \\
        Phi 3 small (7B)        & {8.0} & 14.0 &  9.0 & 4.8 & 14.0 & 6.3 & 7.7 \\
        \midrule
        
        \rowcolor[rgb]{0.9,0.9,0.9} EXAONE 3.5 2.4B & \textbf{80.8} & \textbf{100.0} & \textbf{84.0} & \textbf{61.6} & {84.0} & \textbf{65.3} & \textbf{74.7} \\
        Qwen 2.5 3B     & \underline{56.4} & \underline{98.0} & \underline{63.3} & 2.4 & \textbf{94.0} & {17.7} & \underline{40.5} \\
        Qwen 2.5 1.5B   & {22.0} & 96.0 & 34.3 & 21.6 & \underline{92.0} & 33.3 & 33.8 \\
        Llama 3.2 3B    & 48.8 & {12.0} & {42.7} & \underline{40.0} & {16.0} & \underline{36.0} & {39.3} \\
        Gemma 2 2B      & - & - & - & - & - & - & - \\
        \bottomrule
    \end{tabular}
    \vspace{2mm}
    \caption{Performance comparison results of EXAONE 3.5 language models with similar-sized recently released language models with Ko-LongRAG benchmarks. The benchmark is extended with the ``Unanswerable'' case, which requires models to respond as ``Unanswerable'' when the information cannot be found within the context. \textbf{Bold} scores indicate the best performance, and \underline{underlined} scores mean the second best.}
    \label{tab:kolongrag_details}
\end{table}

The detailed task scores are presented in Table~\ref{tab:kolongrag_details}.  Similar to \textsc{LongRAG}, the evaluation setup for \textsc{Ko-LongRAG} incorporates short answer evaluation to align with the methodology used in \textsc{LongRAG}. However, as the trends for short answers are consistent with those observed for long answers, the final evaluation focuses solely on long answer correctness to streamline the analysis without compromising robustness. 

The detailed prompt and examples used for \textsc{Ko-LongRAG} evaluation are provided in Figure~\ref{fig:Ko-LongRAG-Prompt}~and~\ref{fig:Ko-LongRAG-Examples}, respectively. The prompt used for \textsc{Ko-LongRAG} evaluation is illustrated in Figure~\ref{fig:Ko-LongRAG-Judge-prompt}.
\clearpage
\begin{figure}[H]
\centering
\begin{tcolorbox}[
  title=Ko-LongRAG Prompt, % 제목
  colframe=Black!80!White,   % 테두리 색상
  colback=gray!10,          % 배경 색상
  coltitle=white,           % 제목 텍스트 색상
  colbacktitle=Black!80!White, % 제목 배경 색상
  fonttitle=\bfseries,      % 제목의 글꼴 스타일
  breakable=false,          % breakable 비활성화
  rounded corners,          % 둥근 모서리
  boxsep=3pt,               % 박스 안 여백
  width=\textwidth          % 박스 폭을 페이지 폭으로 설정
]

\textbf{[ Single-doc QA Prompt ]} \\  
\\
\textbf{System:} 당신은 도움이 되는 어시스턴트입니다.\\
\textbf{User:} \\
다음 문서를 살펴보고, 질문에 대한 답을 추출하세요. \\
질문에 대한 답만 생성하세요. 답변은 매우 간결해야 합니다. \\
답변을 문서에서 찾을 수 없는 경우, \\
'주어진 정보로 답할 수 없다'로 응답하세요. \\

문서는 “Title”에 따라 정렬된 Wikipedia 문단 목록이며 제목은 다음과 같습니다: \{\{\textsl{titles}\}\}. \\
각 위키피디아 문단은 'Title' 필드와 'Text' 필드를 포함합니다. \\
문서는 다음과 같습니다: \{\{\textsl{context}\}\}. 질문은 다음과 같습니다: \{\{\textsl{question}\}\}. \\

\vspace{0.5cm}

\textbf{[ Multi-doc QA Prompt ]} \\  
\\
\textbf{System:} 당신은 도움이 되는 어시스턴트입니다.\\
\textbf{User:} \\
다음 문서를 검토하고 질문에 답하세요. \\
문서는 Wikipedia 문단 목록이며 제목은 다음과 같습니다: \{\{\textsl{titles}\}\}. \\
질문에는 두 가지 유형이 있습니다: \\
예 또는 아니오로 답하거나 두 후보 중에서 선택해야 하는 비교 질문과, 단답형 형태의 일반 질문입니다. \\
문서는 다음과 같습니다: \{\{\textsl{context}\}\}. \\
문서에서 필요한 문단을 찾아 질문에 답하세요: \{\{\textsl{question}\}\}. \\
일반 질문의 경우 문단에서 정확한 단어를 찾아서 답변해야 합니다. \\
질문에 대한 답만 생성하고 다른 어떤 것도 생성하지 마세요. \\
답변을 문서에서 찾을 수 없는 경우, '주어진 정보로 답할 수 없다'로 응답하세요. \\

\end{tcolorbox}
\caption{Prompt for evaluating Ko-LongRAG.}
\label{fig:Ko-LongRAG-Prompt}
\end{figure}

\begin{figure}[p]
\centering
\begin{tcolorbox}[
  title=Ko-LongRAG Examples, % 제목
  colframe=Black!80!White,   % 테두리 색상
  colback=gray!10,          % 배경 색상
  coltitle=white,           % 제목 텍스트 색상
  colbacktitle=Black!80!White, % 제목 배경 색상
  fonttitle=\bfseries,      % 제목의 글꼴 스타일
  breakable=false,          % breakable 비활성화
  rounded corners,          % 둥근 모서리
  boxsep=5pt,               % 박스 안 여백
  width=\textwidth          % 박스 폭을 페이지 폭으로 설정
]

\textbf{[ Single-doc QA Answerable Case ]} \\
\\
\textbf{Context:} \\  
... \\  
Title: 박진우 (야구인) \\
Text: 박진우(朴晋佑, 1990년 2월 12일 $\sim$ )는 전 KBO 리그 NC 다이노스의 투수이자, 현 KBO 리그 SSG 랜더스의 스카우트이다. \\  
...\\
2019년 시즌 : 선발과 불펜을 가리지 않고 활약했다. 시즌 140.2이닝 3점대 평균자책점, 92탈삼진, 9승 7패, 5홀드를 기록했다. 이동욱 감독은 '가장 MVP로 꼽고 싶은 선수'라며 칭찬했다.\\
...\\
\\
\textbf{Question:} 박진우가 NC 다이노스에서 9승을 기록한 시즌은 언제인가요? \\
\textbf{Answer:} 2019년 \\

\vspace{0.3cm}

\textbf{[ Single-doc QA Unanswerable Case ]} \\
\\
\textbf{Question:} 인천남동소방서의 설립 연도는 무엇인가요? \\
\textbf{Answer:} 주어진 문서내에서 답할 수 있는 정보가 충분하지 않습니다. \\

\vspace{0.5cm}

\textbf{[ Multi-doc QA Answerable Case ]} \\
\\
\textbf{Context:} \\  
...\\
Title: 아스투리아스 공상\\
Text: 아스투리아스 공상은 스페인의 프린시페 데 아스투리아스 재단(Fundación Príncipe de Asturias)이 주관하는 상이다. 1980년 9월 24일 스페인의 왕세자에 해당하는 호칭인 아스투리아스 공이었던 펠리페 (Felipe, 펠리페 6세)에 의해 제정되었으며 1981년에 첫 시상식이 열렸다. 총 9개 부문 (예술 부문, 커뮤니케이션·인문주의 부문, 국제 협력 부문, 문학 부문, 사회과학 부문, 체육 부문, 기술·과학 연구 부문, 화합 부문, 아스투리아스 모범상 부문)으로 나누어 시상한다. 시상식은 아스투리아스 지방의 오비에도에서 열린다. 수상자는 주안 미로가 제작한 조각, 상금 50,000 유로를 받게 된다.\\
...\\
Title: 미겔 데 세르반테스 상 \\
Text: 미겔 데 세르반테스 상(-賞, ) 또는 세르반테스 상은 스페인 작가 미겔 데 세르반테스의 이름이 붙은 스페인어 작가에게 수여되는 문학상으로, 영연방의 맨 부커 상과 유사한 스페인어권의 상이다. 그러나 맨 부커 상과는 다르게 일생 동안의 문학적 성취를 평가해서 단 한 번만 수여하므로 스페인어권에서 그 권위는 노벨 문학상에 버금간다. 1976년 제정되었다. 스페인 문화부가 수여하며 상금은 12만 5천 유로이다.\\
...\\
\textbf{Question:} 아스투리아스 공상과 미겔 데 세르반테스 상 중 상금이 더 많은 것은 무엇인가요? \\
\textbf{Answer:} 미겔 데 세르반테스 상 \\

\vspace{0.3cm}

\textbf{[ Multi-doc QA Unanswerable Case ]} \\
\\
\textbf{Question:} 넬슨 록펠러와 노아 사이러스는 둘 다 정치 경력을 가지고 있었나요? \\
\textbf{Answer:} 주어진 문서내에서 답할 수 있는 정보가 충분하지 않습니다. \\

\end{tcolorbox}
\caption{Examples of Ko-LongRAG.}
\label{fig:Ko-LongRAG-Examples}
\end{figure}

\begin{figure}[p]
\centering
\begin{tcolorbox}[
  title=Ko-LongRAG LLM-as-a-Judge Prompt, % 제목
  colframe=Black!80!White,   % 테두리 색상
  colback=gray!10,          % 배경 색상
  coltitle=white,           % 제목 텍스트 색상
  colbacktitle=Black!80!White, % 제목 배경 색상
  fonttitle=\bfseries,      % 제목의 글꼴 스타일
  breakable=false,          % 내용이 길어도 자동 분리 비활성화
  rounded corners,          % 둥근 모서리
  width=\textwidth          % 박스 폭을 페이지 폭으로 설정
]

\textbf{System:} \\  

You are an expert evaluator of text answers in Korean. \\  
Your task is to compare the content of two Korean answers, a long answer (\texttt{long\_ans}) and a short answer (\texttt{short\_ans}), with the provided correct answers (\texttt{Answer}), which may contain multiple correct options. \\  
Both the long answer and the short answer need to be checked for correctness. The long and short answers do not need to match any of the answers in the \texttt{Answer} list word-for-word but must convey the same key meaning or idea. \\  
If either the long or short answer matches any one of the correct answers in the \texttt{Answer} list, it should be considered correct. \\  
Focus only on the accuracy of the content and ignore style, tone, or extra information unless it introduces inaccuracies. \\  
For both the long and short answers, return only the evaluation result as a Python dictionary object, and ensure the output is formatted as valid Python code. \\  

Here are two examples of how to evaluate answers: \\  

Example 1: \\  
\texttt{Question}: HP는 게임에서 무엇을 의미하나요? \\  
\texttt{Answer}: [\textquotesingle 체력\textquotesingle , \textquotesingle 생명력\textquotesingle ] \\  
\texttt{long\_ans}: HP는 '생명력' 또는 '체력'을 의미하며, 게임에서 캐릭터의 생존력을 나타내는 지표입니다. HP가 줄어들면 캐릭터는 점점 약해지며, 0이 되면 게임에서 탈락하거나 패배할 수 있습니다. \\  
\texttt{short\_ans}: HP는 캐릭터의 체력입니다. \\  
\texttt{Evaluation}: \{\textquotesingle long\_ans\textquotesingle : \textquotesingle correct\textquotesingle , \textquotesingle short\_ans\textquotesingle : \textquotesingle correct\textquotesingle \} \\  

Example 2: \\  
\texttt{Question}: 프랑스의 수도는 어디인가요? \\  
\texttt{Answer}: [\textquotesingle 파리\textquotesingle ] \\  
\texttt{long\_ans}: 프랑스의 수도는 파리로, 리옹의 오른쪽 아래에 위치하고, 문화와 예술의 중심지로 알려져 있습니다. 에펠탑, 루브르 박물관, 노트르담 대성당 등 유명한 관광지가 위치해 있습니다. \\  
\texttt{short\_ans}: 프랑스의 수도는 리옹입니다. \\  
\texttt{Evaluation}: \{\textquotesingle  long\_ans\textquotesingle : \textquotesingle correct\textquotesingle , \textquotesingle short\_ans\textquotesingle : \textquotesingle incorrect\textquotesingle \} \\  

Now, proceed with your evaluation of the following question, answer, and responses, and return only the evaluation as a valid Python dictionary. \\  
Ensure the response is a valid Python dictionary object without any additional text. \\  

\vspace{0.5cm}

\textbf{User:} \\  

Evaluate the following long and short answers based on the provided correct answer. \\  
Your goal is to determine if the long and short answers are correct. \\  
Return the evaluation result in the form of a Python dictionary: \texttt{\{\textquotesingle long\_ans\textquotesingle: \textquotesingle correct \textquotesingle or \textquotesingle incorrect \textquotesingle,  \textquotesingle short\_ans\textquotesingle:  \textquotesingle correct\textquotesingle or  \textquotesingle incorrect\textquotesingle}\}. \\  

\texttt{Question}: \{\{\textsl{question}\}\} \\  
\texttt{Answer}: \{\{\textsl{answer}\}\} \\  
\texttt{long\_ans}: \{\{\textsl{long\_ans}\}\} \\  
\texttt{short\_ans}: \{\{\textsl{short\_ans}\}\} \\  

Return only the evaluation in the form of a Python dictionary. \\  
Do not include any explanation or additional comments. \\  

\end{tcolorbox}
\caption{LLM-as-a-judge prompt for evaluating Ko-LongRAG.}
\label{fig:Ko-LongRAG-Judge-prompt}
\end{figure}

\subsubsection{Ko-WebRAG}
\textsc{Ko-WebRAG} is a real-world benchmark tailored to assess the performance of language models as generators within the Retrieval-Augmented Generation (RAG) framework, using a web-search engine as a fixed retriever. The benchmark comprises 300 RAG tasks, each featuring a user query alongside documents retrieved by the simulated web-search engine. The retrieved documents in \textsc{Ko-WebRAG} are meticulously curated to ensure they provide sufficient supporting information for generating a gold-standard answer. Context lengths vary from 4K to 32K tokens, with an average length of approximately 14,000 tokens. 

Each dataset instance includes a user query, a gold-standard answer, and the corresponding retrieved documents. The performance of the target LLM is evaluated based on its ability to generate answers that match the gold-standard answer, measuring its effectiveness as a generator in RAG tasks.

The evaluation involves assessing the responses of the LLM to each of the 300 tasks using GPT-4o. The percentage of tasks for which the LLM’s responses pass this evaluation is reported as the final score. To align with the purpose of a Korean-language benchmark, the GPT-4o LLM-as-a-judge prompt incorporates additional criteria beyond semantic alignment with the gold-standard answer; it also checks whether questions asked in Korean have been answered in Korean. Note that Qwen models smaller than 32B often fail to meet this language compliance criterion, leading to lower scores.
\clearpage
\subsection{General Domain}\label{appendix:general_dom_setting}

For all benchmarks in the General Domain category, we use 0-shot prompts and parse a final answer from the generated model response. Greedy decoding is used and maximum length of generation is set to 2,048 for all tasks. From Figure~\ref{fig:general_dom_math_prompt}~to~\ref{fig:general_dom_arc_prompt}, we present all prompts we use for the evaluation for each benchmarks. For BBH, we utilize 0-shot CoT prompts\footnote{\url{https://github.com/EleutherAI/lm-evaluation-harness/tree/main/lm_eval/tasks/bbh/cot_zeroshot}} from Language Model Evaluation Harness~\cite{eval-harness}.

\begin{figure}[H]
\centering
\begin{tcolorbox}[
  title=GSM8K/MATH prompt (CoT), % 제목
  colframe=Black!80!White,   % 테두리 색상
  colback=gray!10,          % 배경 색상
  coltitle=white,           % 제목 텍스트 색상
  colbacktitle=Black!80!White, % 제목 배경 색상
  fonttitle=\bfseries,      % 제목의 글꼴 스타일
  breakable=false,          % breakable 비활성화
  rounded corners,          % 둥근 모서리
  boxsep=3pt,               % 박스 안 여백
  width=\textwidth          % 박스 폭을 페이지 폭으로 설정
]

Given the following math problem, reason step-by-step and give a final answer to the problem. Put your final answer within \textbackslash boxed\{\}.\\
Problem: \{\{\textsl{question}\}\}\\
\\
Answer: Let's think step by step.

\end{tcolorbox}
\caption{Prompt for evaluating \textsc{GSM8K} (CoT) and \textsc{MATH} (CoT) benchmarks.}
\label{fig:general_dom_math_prompt}
\end{figure}

\begin{figure}[H]
\centering
\begin{tcolorbox}[
  title=HumanEval/MBPP prompt, % 제목
  colframe=Black!80!White,   % 테두리 색상
  colback=gray!10,          % 배경 색상
  coltitle=white,           % 제목 텍스트 색상
  colbacktitle=Black!80!White, % 제목 배경 색상
  fonttitle=\bfseries,      % 제목의 글꼴 스타일
  breakable=false,          % breakable 비활성화
  rounded corners,          % 둥근 모서리
  boxsep=3pt,               % 박스 안 여백
  width=\textwidth          % 박스 폭을 페이지 폭으로 설정
]

\textbf{User:} \\  

Please provide a self-contained Python script that solves the following problem in a markdown code block:\\
\texttt{```}\\
\{\{\textsl{input}\}\}\\
\texttt{```}\\

\vspace{0.5cm}

\textbf{Assistant:} \\  

Below is a Python script with a self-contained function that solves the problem and passes corresponding tests:\\
\texttt{```}python\\

\end{tcolorbox}
\caption{Prompt for evaluating \textsc{HumanEval} and \textsc{MBPP} benchmarks. We use the default prompt setting from the official Github repository\tablefootnote{\url{https://github.com/evalplus/evalplus}} of EvalPlus~\cite{liu2023is}.}
\label{fig:general_dom_coding_prompt}
\end{figure}

\begin{figure}[H]
\centering
\begin{tcolorbox}[
  title=MMLU/GPQA prompt (CoT), % 제목
  colframe=Black!80!White,   % 테두리 색상
  colback=gray!10,          % 배경 색상
  coltitle=white,           % 제목 텍스트 색상
  colbacktitle=Black!80!White, % 제목 배경 색상
  fonttitle=\bfseries,      % 제목의 글꼴 스타일
  breakable=false,          % breakable 비활성화
  rounded corners,          % 둥근 모서리
  boxsep=3pt,               % 박스 안 여백
  width=\textwidth          % 박스 폭을 페이지 폭으로 설정
]

Given the following question and candidate answers (A, B, C and D), reason step-by-step and choose the best answer to the question.\\
Question: \{\{\textsl{question}\}\}\\
A. \{\{\textsl{option A}\}\}\\
B. \{\{\textsl{option B}\}\}\\
C. \{\{\textsl{option C}\}\}\\
D. \{\{\textsl{option D}\}\}\\
Your response should end with "The best answer is [the\_answer\_letter]" where the [the\_answer\_letter] is one of A, B, C or D.\\
\\
Answer: Let's think step by step.

\end{tcolorbox}
\caption{Prompt for evaluating \textsc{MMLU} (CoT) and \textsc{GPQA} (CoT) benchmarks.}
\label{fig:general_dom_mcqa_cot_prompt}
\end{figure}

\begin{figure}[H]
\centering
\begin{tcolorbox}[
  title=KMMLU prompt (CoT), % 제목
  colframe=Black!80!White,   % 테두리 색상
  colback=gray!10,          % 배경 색상
  coltitle=white,           % 제목 텍스트 색상
  colbacktitle=Black!80!White, % 제목 배경 색상
  fonttitle=\bfseries,      % 제목의 글꼴 스타일
  breakable=false,          % breakable 비활성화
  rounded corners,          % 둥근 모서리
  boxsep=3pt,               % 박스 안 여백
  width=\textwidth          % 박스 폭을 페이지 폭으로 설정
]

다음 시험 문제에 대해서, 충분히 생각하고 추론하여, 4개의 보기(A, B, C, D) 중 정답을 고르세요.\\
문제: \{\{\textsl{question}\}\}\\
A. \{\{\textsl{option A}\}\}\\
B. \{\{\textsl{option B}\}\}\\
C. \{\{\textsl{option C}\}\}\\
D. \{\{\textsl{option D}\}\}\\
당신의 대답은 "정답은 [정답 보기]입니다."로 끝나야하고, [정답 보기]는 A, B, C, D 중 하나여야 합니다.\\
\\
정답: 문제를 풀기 위해, 한 번 천천히 생각해봅시다.

\end{tcolorbox}
\caption{Prompt for evaluating \textsc{KMMLU} (CoT) benchmark.}
\label{fig:general_dom_kmmlu_prompt}
\end{figure}

\begin{figure}[H]
\centering
\begin{tcolorbox}[
  title=ARC-C prompt, % 제목
  colframe=Black!80!White,   % 테두리 색상
  colback=gray!10,          % 배경 색상
  coltitle=white,           % 제목 텍스트 색상
  colbacktitle=Black!80!White, % 제목 배경 색상
  fonttitle=\bfseries,      % 제목의 글꼴 스타일
  breakable=false,          % breakable 비활성화
  rounded corners,          % 둥근 모서리
  boxsep=3pt,               % 박스 안 여백
  width=\textwidth          % 박스 폭을 페이지 폭으로 설정
]

Given the following question and candidate answers (A, B, C and D), choose the best answer to the question.\\
Question: \{\{\textsl{question}\}\}\\
A. \{\{\textsl{option A}\}\}\\
B. \{\{\textsl{option B}\}\}\\
C. \{\{\textsl{option C}\}\}\\
D. \{\{\textsl{option D}\}\}\\
Your response should end with "The best answer is [the\_answer\_letter]" where the [the\_answer\_letter] is one of A, B, C or D.\\
\\
Answer:

\end{tcolorbox}
\caption{Prompt for evaluating \textsc{ARC-C} benchmark.}
\label{fig:general_dom_arc_prompt}
\end{figure}

\newpage
\bibliographystyle{plain} % We choose the "plain" reference style
\bibliography{refs} % Entries are in the refs.bib file

%%%%%%%%%%%%%%%%%%%%%%%%%%%%%%%%%%%%%%%%%%%%%%%%%%%%%%%%%%%%

\end{document}